\documentclass[10pt,twocolumn,letterpaper]{article}

\usepackage{iccv}
\usepackage{times}
\usepackage{epsfig}
\usepackage{graphicx}
\usepackage{amsmath}
\usepackage{amssymb}

\usepackage[accsupp]{axessibility}  % Improves PDF readability for those with disabilities.

\usepackage{xr}
\usepackage{epsfig}
\usepackage{graphicx,comment}
\usepackage{amsmath}

\usepackage{indentfirst}
\usepackage{algorithm}
\usepackage{algpseudocode} 
\usepackage{amssymb}
\usepackage{verbatim}
\usepackage{multirow}
\usepackage{booktabs}
\usepackage{longtable}
\usepackage{soul}
\usepackage{arydshln}
\usepackage{enumitem}
\usepackage[usenames,dvipsnames]{xcolor}
\usepackage{float}
\usepackage{xspace}
\usepackage{bm}
\usepackage[titletoc]{appendix}
\usepackage[resetlabels]{multibib}
\usepackage{capt-of}% or \usepackage{caption}
\newcites{sec}{References}
\newcommand{\SystemName}{QAD}
\newcommand\mfont{\fontsize{7.6pt}{18pt}\selectfont}
\newcommand{\ImageNet}{I{\mfont MAGE}N{\mfont ET}}
\newcommand{\ResNet}{R{\mfont ES}N{\mfont ET}}
\newcommand{\EffNet}{E{\mfont FFICIENT}N{\mfont ET}}
\usepackage{upgreek}

\newcommand\Tau{\scalebox{1.44}{$\tau$}}
\usepackage{color, colortbl}
\definecolor{codegreen}{rgb}{0,0.6,0}
\definecolor{codegray}{rgb}{0.5,0.5,0.5}
\definecolor{codepurple}{rgb}{0.58,0,0.82}
\definecolor{backcolour}{RGB}{230, 230, 230}
\definecolor{LightCyan}{rgb}{0.88,1,1}
\definecolor{LightRed}{RGB}{255, 204, 203}
\usepackage{amsmath}
\DeclareMathOperator*{\argmax}{arg\,max}

\newcommand\Tstrut{\rule{0pt}{2.6ex}}       % "top" strut
\newcommand\Bstrut{\rule[-1.1ex]{0pt}{0pt}} % "bottom" strut
\newcommand\VRule[1][\arrayrulewidth]{\vrule width #1}
\makeatletter
\def\adl@drawiv#1#2#3{%
        \hskip.5\tabcolsep
        \xleaders#3{#2.5\@tempdimb #1{1}#2.5\@tempdimb}%
                #2\z@ plus1fil minus1fil\relax
        \hskip.5\tabcolsep}
\newcommand{\cdashlinelr}[1]{%
  \noalign{\vskip\aboverulesep
           \global\let\@dashdrawstore\adl@draw
           \global\let\adl@draw\adl@drawiv}
  \cdashline{#1}
  \noalign{\global\let\adl@draw\@dashdrawstore
           \vskip\belowrulesep}}
\makeatother

\usepackage{pifont}% http://ctan.org/pkg/pifont
\newcommand{\xmark}{\ding{55}}%
\usepackage[pagebackref=true,breaklinks=true,letterpaper=true,colorlinks,bookmarks=false]{hyperref}

\iccvfinalcopy % *** Uncomment this line for the final submission

\ificcvfinal\fi

\begin{document}

%%%%%%%%% TITLE
\title{Quality-Agnostic Deepfake Detection with Intra-model Collaborative Learning}

\author{Binh M. Le \\
Dept. of Computer Science \& Engineering\\
Sungkyunkwan University\\ Suwon, South Korea\\
{\tt\small bmle@g.skku.edu}
\and
Simon S. Woo\thanks{Corresponding author.}  \\
Dept. of Computer Science \& Engineering\\
Sungkyunkwan University\\ Suwon, South Korea\\
{\tt\small swoo@g.skku.edu}
}

\maketitle
\ificcvfinal\thispagestyle{empty}\fi

%%%%%%%%% ABSTRACT
\begin{abstract}
Deepfake has recently raised a plethora of societal concerns over its possible security threats and dissemination of fake information. Much research on deepfake detection has been undertaken. However, detecting low quality as well as simultaneously detecting different qualities of deepfakes still remains a grave challenge. Most SOTA approaches are limited by using a single {specific} model for {detecting certain deepfake video quality type}. When constructing multiple models with prior information about video quality, this kind of strategy incurs significant computational cost, as well as model and training data overhead. {Further, it cannot be scalable and practical to deploy in real-world settings.} In this work, we propose a universal intra-model collaborative learning framework to enable the {effective and simultaneous} detection of different quality of deepfakes. That is, our approach is the quality-agnostic deepfake detection method, dubbed \SystemName. In particular, by observing the upper bound of general error expectation, we maximize the {dependency} between intermediate representations of images from different quality levels via \textit{Hilbert-Schmidt Independence Criterion}. In addition, an \textit{Adversarial Weight Perturbation} module is carefully devised to enable the model to be more robust against image corruption while boosting the overall model's performance. Extensive experiments over seven popular deepfake datasets demonstrate the superiority of our \SystemName\ model over prior SOTA benchmarks.
\end{abstract}

%%%%%%%%% BODY TEXT
\section{Introduction}

Deep learning approaches for facial manipulation, such as deepfakes, have recently received {considerable attention} \cite{wang2019fakespotter,li2020face, zhao2021multi, khayatkhoei2020spatial, hu2022finfer, jeong2022frepgan}, because they can be abused for the malicious purposes such as fake news, pornography, etc. Due to the advancements made in Generative Adversarial Networks and other deep learning-based computer vision algorithms, deepfakes have also become more realistic and natural, making it harder not only for humans, but also for classifiers to tell them apart. Moreover, it has been simpler than {ever before} to create convincing deepfakes {using simple programs and apps without requiring advanced machine learning knowledge}. Such easy-to-create and realistic fake images and videos can be maliciously exploited, raising significant security, privacy, and societal concerns such as fake news propagation \cite{quandt2019fake}, and stealing personal information via {phishing and scams}~\cite{fbi2022}.% which is recently reported by FBI. 

\begin{figure}[t]
\centering
\includegraphics[width=8.5cm]{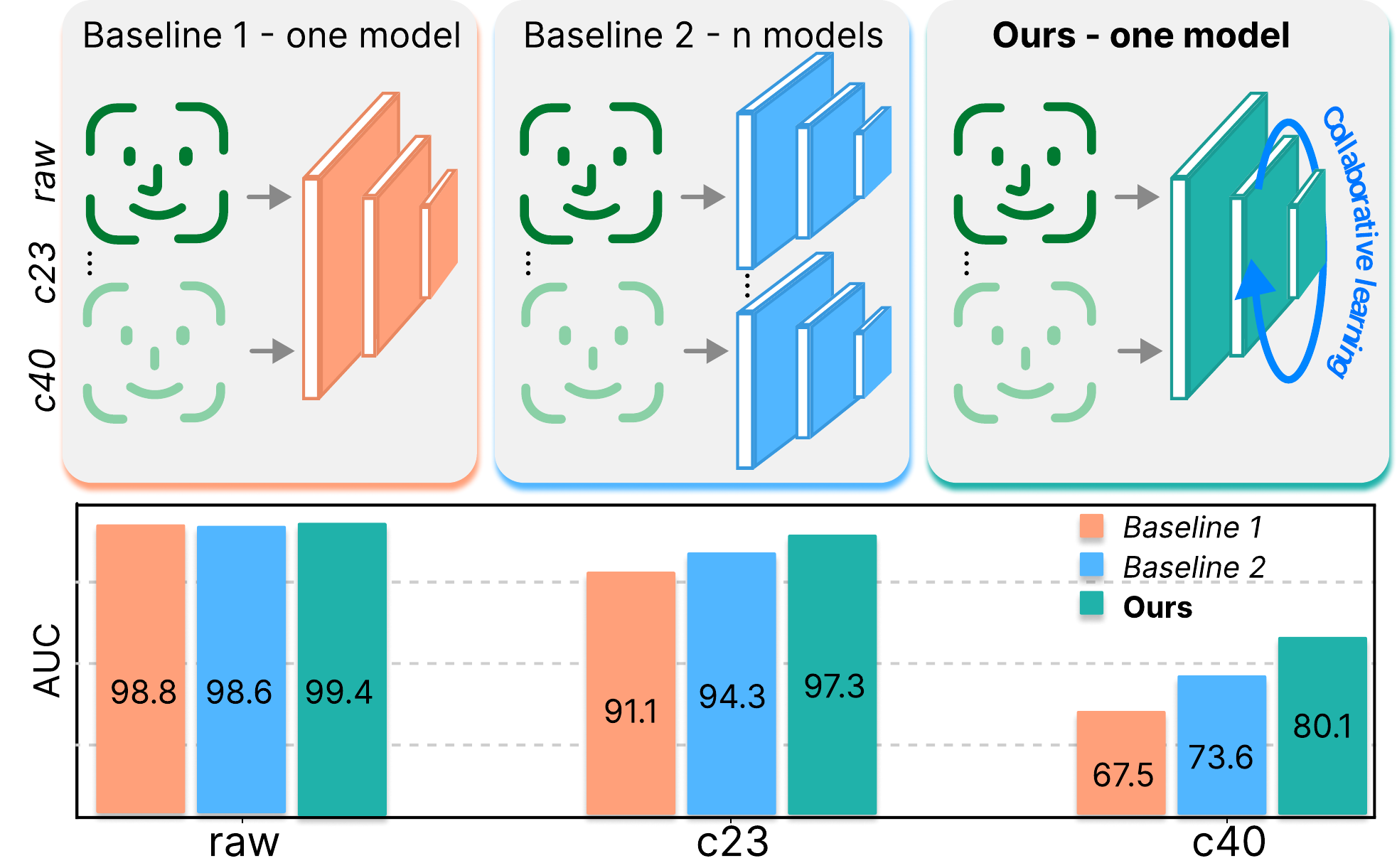}
\caption{\small \textbf{A summary of our goal.} Our approach stands out from previous works that detect deepfakes using separate models for different qualities (\textit{e.g.} Baseline 2 \cite{ble2022add}) or a single model without considering the interaction between qualities (\textit{e.g.} Baseline 1 \cite{rossler2019faceforensics++}). Instead, our method employs all quality levels and improves the performance of the model on each quality level, leading to overall enhanced performance.
}
\label{fig:motivate}
\vspace{-8pt}
\end{figure}

To mitigate such problems caused by deepfakes, there has been a tremendous research effort put into constructing reliable detectors~\cite{li2018exposing, khayatkhoei2020spatial, dzanic2019fourier, qian2020thinking, zhao2021multi, sebyakin2021spatio}. Although they have achieved outstanding performance with high-quality deepfakes, most of them have failed to detect low-quality deepfakes effectively ~\cite{dzanic2019fourier, rossler2019faceforensics++}. While video compression steps do not {significantly} impact on visualization, it drastically drop deepfake detectors' {performance on low-quality deepfakes (c40)}. A handful of research has been focused on detecting low-quality deepfakes such as ADD~\cite{ble2022add} and BZNet \cite{lee2022bznet}. However, their methods can only detect low-quality compressed deepfakes. And, those prior approaches expose a critical problem, when deployed in practice since the prior video quality information of the input is unknown. Moreover, developing different models for each input quality induces significant computational overhead. Other works, such as LipForensics \cite{haliassos2021lips}, also attempt to make their detectors robust against various corruptions and compression. Nevertheless, it is unable to detect image-based deepfakes with random lossy compression like JPEG. 

In this research, we propose the novel deepfake detection method, \SystemName, which can simultaneously detect both high and low-quality (quality-agnostic) deepfakes in a single model, as illustrated in Fig. \ref{fig:motivate}. Especially, we propose a universal intra-model collaborative learning framework to provide the effective detection of different quality deepfakes. We modulate the conventional model-based collaborative learning \cite{song2018collaborative} to an instance-based intra-model collaborative learning framework in our training. During the training phase, our single model simultaneously learns the representations of one image, but with different qualities. By utilizing the collaborative learning framework, our~\SystemName\ can align the distributions of high and low-quality image representations to be geometrically similar. Hence, it can avoid the overfitting caused by compressed images and the overconfidence caused by raw images, {while} boosting its overall performance. 

In particular, {we perform a rigorous} theoretical analysis, and show that the low-quality deepfake classification error can be bounded by two terms: classification loss and the distance between the representations of high and low-quality images. Instead of using a direct pairwise regularization to minimize the gaps between the high and low-quality image representations, we propose to apply \textit{Hilbert-Schmidt Independence Criterion} (HSIC) to maximize the dependence between a mini-batch of high and low-quality images, thus maximizing the mutual information between them, and supporting the high-level representations and effective output predictions. Meanwhile, to enhance the model's robustness under heavy input compression, we propose \textit{Adversarial Weight Perturbation} (AWP)~\cite{wu2020adversarial, calian2021defending}, which can further flatten the weight loss landscape of the model, bridging the gap in multiple quality learning for deepfake detection.
 
 Finally, we conduct extensive experiments to show the effectiveness of our~\SystemName\ with seven different popular benchmark datasets. We first show that our method can outperform previous baselines when training with data from various video and image compression qualities. Furthermore, we show that our~\SystemName\ exceeds the performance of {the  SOTA} quality-aware models such as BZNet \cite{lee2022bznet} by a significant margin, while requiring remarkably fewer computational parameters and no prior knowledge of the inputs. Our contributions are summarized as follows:
 % \footnote{Our code is available at \url{https://anonymous.4open.science/r/QAD-5523}, and we plan to release the entire code upon acceptance of the paper.}

\noindent \textbf{1)}  We theoretically analyze and prove that the classification error of low-quality deepfakes can be bounded by its classification loss and the representation distance with its corresponding high-quality images.

\noindent \textbf{2)}  We propose a unified quality-agnostic deepfake detection framework (\SystemName), utilizing instance-based intra-model collaborative learning. We use the \textit{Hilbert-Schmidt Independence Criterion} (HSIC) to maximize the geometrical similarity between intermediate representations of high and low-quality deepfakes, and \textit{Adversarial Weight Perturbation} (AWP) to make our model robust under varying input compression.

\noindent \textbf{3)} We demonstrate that our approach outperforms well-known baselines, including the total of \textit{eight} quality-agnostic and quality-aware SOTA methods with \textit{seven} popular benchmark datasets.

\section{Related works}
\subsection{Deepfake detection} Recently, deepfakes have been of the utmost crucial because they can cause serious security and privacy threats. Threrefore, a large number of detection methods have been proposed to effectively identify such deepfakes \cite{rossler2019faceforensics++, li2020face,li2019faceshifter,jeon2020t, rahmouni2017distinguishing, wang2019fakespotter, li2018exposing, khayatkhoei2020spatial, dzanic2019fourier, zhao2021multi}. However, the majority of the aforementioned works focus on mining visual artifacts of deepfakes, such as the blending boundaries of generated faces \cite{li2020face}, the irregularity of pupil shapes \cite{guo2022eyes}, the spatiotemporal inconsistency \cite{de2020deepfake,sebyakin2021spatio}, or exploring deep learning-based attention methods \cite{zhao2021multi} to identify such artifacts. Meanwhile, several approaches also showed that exposing deepfakes in the frequency domain is effective. Such methods include analyzing the discrepancies of frequency spectrum  \cite{dzanic2019fourier,binh2021exploring, khayatkhoei2020spatial}, employing the checkerboard artifacts caused by the transposed convolutional operator \cite{zhang2019detecting, frank2020leveraging}, or mining the statistical frequency features with dual deep learning models \cite{qian2020thinking}. Nevertheless, such models' performance substantially decreases when encountering low-quality compressed images. To remedy the above shortcoming, recent studies proposed methods to detect the deepfake in highly compressed low-quality versions such as \cite{ble2022add}, which utilized a knowledge distillation. Also, \cite{lee2022bznet} presented a supper-resolution-based network for enhancing the performance of low-quality deepfake detection. However, all of the aforementioned approaches are limited in developing a single model for each quality of deepfakes, which is impractical to deploy in real-world scenarios due to the requirement of prior knowledge of the input quality.

\subsection{Collaborative learning}

Collaborative learning proposed by \cite{song2018collaborative} is designed to achieve a global minimum of a deep neural network, while maintaining the same computational complexity at inference time as at training time. Collaborative learning inherits the advantages of auxiliary training \cite{szegedy2015going}, multi-task learning \cite{yang2017deep}, and knowledge distillation \cite{hinton2015distilling}. Its applications cover supporting weakly-supervised learning \cite{yuanmm21}, or integrating with online knowledge distillation \cite{wu2021peer, guo2020online}. And, its training graph is divided into  two or more sub-networks to ensure global minimum achievement~\cite{song2018collaborative}. 
Besides, \cite{fang2021intra} proposed an intra-model collaborative learning framework that shares a similar characteristic with self-knowledge distillation. However, all of the approaches are model-based collaborative learning, in which a single input generates multiple outputs (or \textit{views}) through multiple classifier heads of one target network in both training and inference phase.

In this work, we distinguish ourselves by deploying the collaborative learning framework for simultaneously training deepfakes of various qualities with an undeviated single model, namely \textit{instance-based collaborative learning}. Different from conventional mini-batch stochastic optimization, which independently samples random images from different qualities and optimizes the detector, {our} collaborative learning {approach} allows us to utilize the common features in the same image but from different qualities simultaneously. Thus, our deepfake detector circumvent the overfitting caused by compressed images or the overconfidence {from} raw images, enhancing its overall performance.

\subsection{Hilbert-Schmidt Independence Criterion}
The Hilbert-Schmidt Independence Criterion (HSIC)  \cite{gretton2005measuring} measures the statistical dependency between probability distributions. In fact, HSIC differs from the covariance, where $Cov(X, Y)=0$ does not imply that \textit{two} random variables $X$ and $Y$ are independent \cite{renyi1959measures}, while HSIC shows its tractable computation and equivalency in terms of the independence property \cite{gretton2005measuring}. Moreover, HISC is easy to {be} estimated statistically and algorithmically. In practice, applications based on HSIC are found in a variety of {practical} domains, including maximizing the dependencies for self-supervised learning \cite{li2021self} and classification learning \cite{ma2020hsic, greenfeld2020robust}, or defense against model inversion attacks \cite{peng2022bilateral}. In this paper, we utilize the HSIC to maximize the  {dependency} between distributions of deepfake images of different qualities at intermediate layers. Therefore, we aim to constrain low-level representations of images not to be exactly the same, but to share a geometrical similarity of learning features that can support high-level output predictions. 
\section{Methods}
In this section, we first theoretically examine the upper bound for our optimization problem by considering a DNN classifier of $K$ classes, and two modalities of input quality: raw ({high-quality}) and compressed ({low-quality}). 
Then, we discuss how to more efficiently collaborate on the representations of deepfake images of differing quality.

\begin{figure*}[t]
\centering
\includegraphics[width=16.5cm]{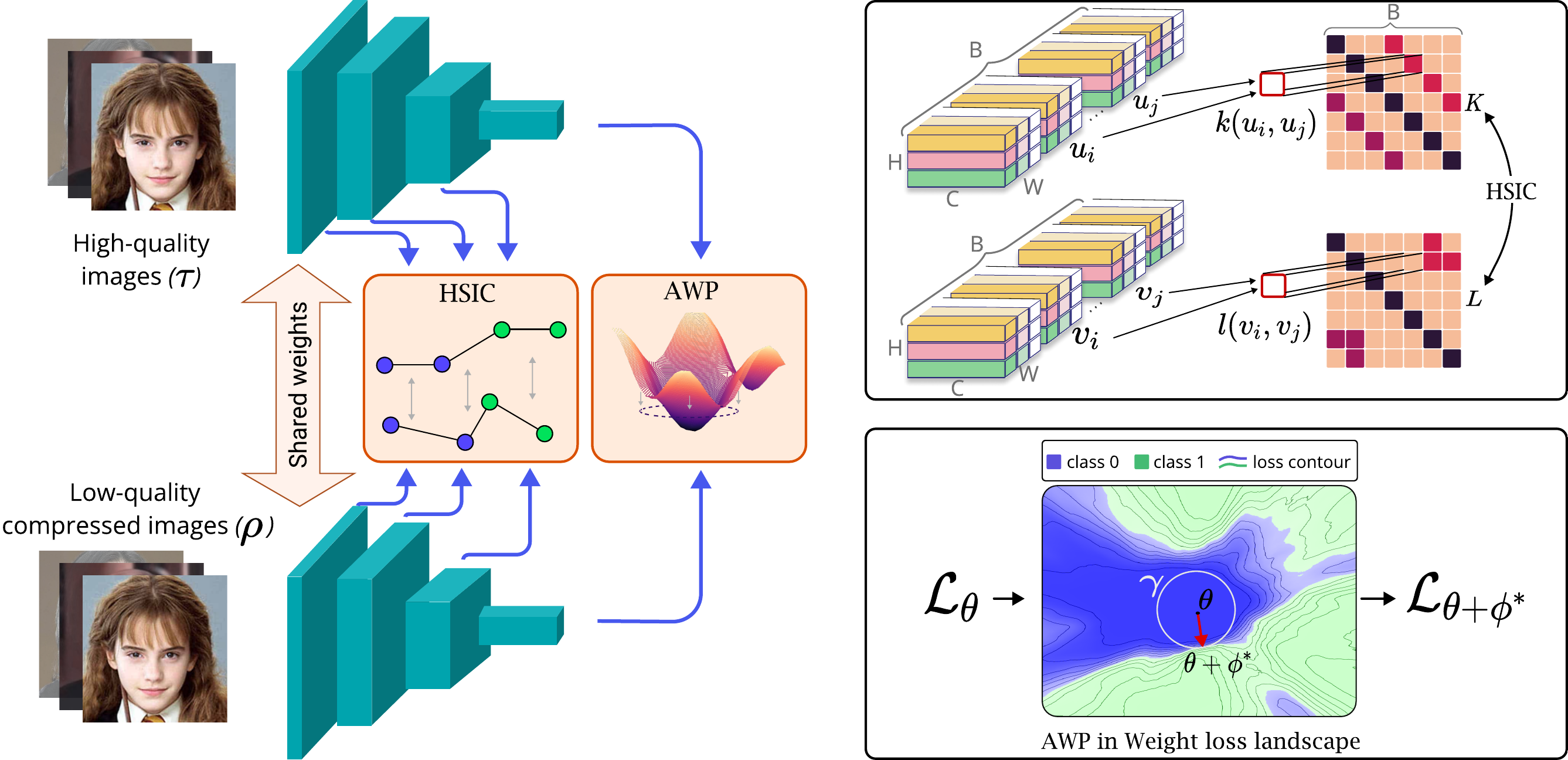}
\caption{\textbf{Overview of our~\SystemName\ framework}. A mini-batch of images from different quality modalities (\textit{e.g.}, two in this diagram) is forwarded through a \textbf{single universal model}. Although it is one model, we pictorially split it into two branches for the reader's understanding. After training, we obtain one universal model that regulates different qualities. \textbf{Top-right}: The HSIC geometrically maximizes the dependency between images from various quality modalities at different resolutions, supporting high-level output predictions. \textbf{Bottom-right}: Through searching for the worst-case parameters' corruption and compensating for input corruption (compression), the AWP flattens the model's weight loss landscape, making the model robust under varying input compression.}
\label{fig:main_arch}
% \vspace{-12pt}
\end{figure*}

\subsection{Preliminary  \& our inspiration}
% Let $\mathcal{X}$ be a data space, in which we endow a data distribution $\mathbb{P}$ with a density function $p(x)$.
Given a  sample $x_r$ from a  space $\mathcal{X}$ and its compressed version at quantile $c$, $x_{c}$ can be expressed as $x_{c} = x_r-\delta_{c}$, and {we define} the corresponding label $y\in \{0,1\}$ (real and fake). Next, a family $\mathcal{F}$ of learning functions $f: \mathcal{X} \rightarrow \mathbb{R}^{2}$ returns a $2$-tuple $f(x)=[f(x, j)]_{j=1}^2$, whose $f(x,j)$ is proportional to the probability to assign $x$ to the {$j$-th} class, and $f$ is  {defined} by learning parameters $\theta$.
% and $\mathcal{Y}_{\Delta}:=\{a \in \mathbb{R}^C :|| a ||_1=1  \wedge a \geq \mathbf{0} \}$ is a $C$-simplex label space. 
Given a training data $\mathcal{D}=\{(x_i, y_i) \}_{i=1}^{n} \subset \mathcal{X} \times \mathcal{Y}$, our goal is to minimize the expectation of a loss function $\mathcal{L}: \mathbb{R}^2 \times \mathcal{Y} \rightarrow \mathbb{R}$. Here, we consider $\mathcal{L}(f(x), y)=1-\sigma_T(f(x,y))$, where $\sigma_T$ is the softmax function with temperature $T > 0$: 
\begin{equation}
    \sigma_T({f(x), y}) = \frac{\text{exp}(f(x,y) / T)}{\sum_{k=1}^{2}\text{exp}(f(x,k)/T)}.
\end{equation}

\textbf{Theorem 1. }\textit{ (Proof can be done similarly as \cite{muhammad2021mixacm} and \cite{hsu2021generalization} in Supp. Material) For any $f \in \mathcal{F}$, and with probability $1-\delta$ over the draw of $\mathcal{D}$,} 
\begin{align}
    \mathbb{E}[\mathbb{I}\{\hat{y}(x_{c}) \neq y\}] &\leq 2\mathbb{E}_{\mathcal{D}}\mathcal{L}(f(x_{c}), y)  \nonumber \\ 
    & + \frac{8}{T}\mathbb{E}_{\mathcal{D}}\mathcal{L}_{i\textendash col}(f(x_r), f(x_c)) + 4 \mathfrak{R}_{\mathcal{D}}(\Phi_{\mathcal{W}}) \nonumber \\ 
    & + \frac{16}{n} + \mathcal{O}\left(\sqrt{\frac{\text{log}(2/\delta)}{2n}}\right), \label{eqn:theorem1}
\end{align}
\textit{where $\mathfrak{R}_{\mathcal{D}}$ is the Rademacher complexity, $\Phi_{\mathcal{W}}= \{\mathcal{L}(f(x_r),y), f\ \in \mathcal{F}\}$, and }
\begin{equation}
    \mathcal{L}_{i\textendash col}(f(x_r), f(x_c))=\parallel f(x_r) - f(x_c) \parallel.
\label{eqn:col}
\end{equation}

\textbf{Insight of Theorem 1. }On the right-hand side of Eq. \ref{eqn:theorem1}, our classifier $f$ depends on two terms, where the first term is the classification loss $\mathcal{L}(f(x_c), y)$ applied to the prediction of the compressed image $x_c$. And, the second term is the instance-based collaborative loss $\mathcal{L}_{i\textendash col}(f(x_r), f(x_c))$ that measures the pairwise difference between predictions of the raw image and its compressed version. Therefore, minimizing the expectation over training data $\mathcal{D}$ of $2\mathcal{L}(f(x_{c}), y) + 8\mathcal{L}_{i\textendash col}(f(x_r), f(x_c))/T$, can decrease the true error. Note that Eq. \ref{eqn:theorem1} is also general so that it can be applicable for raw images. Hence, in practice, the first term can be generalized to both raw and compressed images.

In order to minimize the expectation of errors in both raw and compressed deepfake image predictions, our theoretical analysis shows that we can minimize both classification loss and collaborative loss at the output. However, as observed by \cite{fang2021intra} and in our experiments (see Tab. \ref{tab:ablation_loss}), this instance-based collaborative learning loss fails to achieve the best performance. Additionally, training solely with {highly}-compressed images makes the detector prone to overfitting, yielding a considerable gap between training and test performance \cite{ble2022add}. As a result, the major research challenge is how we can further lower the sensitivity of $f_\theta$ to $x$ at various compression settings and push their representations close to each other.

\textbf{Classification loss.} We can construct a robust model under input corruption by flattening the weight loss landscape. In particular, we apply the \textit{Adversarial Weight Perturbation} \cite{wu2020adversarial} to search for the worst-case perturbations $\phi^{\ast}$ of the model weights at every training step. Thereafter, optimizing the perturbed model via $\mathcal{L}(f_{\theta + \phi^{\ast}}, y)$ can enable it to be more robust under varying input image corruptions/distortions, which represent the varying qualitiies of deepfake inputs (See Fig. \ref{fig:main_arch}\textemdash Bottom-right panel). Furthermore, the classification loss in Eq. \ref{eqn:theorem1} is upper bounded by this new loss function due to the worst-case perturbations.

\textbf{Collaborative loss. }With respect to the collaborative learning loss, Eq. \ref{eqn:col} shares similar characteristics with recent knowledge distillation research \cite{hinton2015distilling, tian2019contrastive}. However, our goal is to {develop and} train a single universal model, not having a teacher-student relationship. Nevertheless, we argue that the gap between raw and compressed image representations can be minimized more efficiently by regularizing their discrepancy at the low-level representations. Moreover, enforcing the similarity of the pairwise representations at the intermediate layers with an input difference of $|\delta_c|$ can collapse layers' weights to zero or can lead a deep model to remember training data instead of learning discriminative features. Therefore, we relax this  {constraint} by maximizing the kernel dependency in a mini-batch of data between the raw and compressed image representations by the \textit{Hilbert-Schmidt Independence Criterion} (HSIC). Using HSIC, we can instead enforce the geometrical structures of mini-batch of raw data and the compressed data to be similar, so that we can still effectively detect different-quality deepfakes in a single model. From a mutual information perspective, maximizing the kernel dependency can enforce the mutual information of the learned representations of different compression ratios, thus regularizing the detector to be more generalized (See Fig. \ref{fig:main_arch}\textemdash Top-right panel).

\subsection{Details of our methods}
\subsubsection{Weight loss landscape flattening}
Recent studies \cite{ishida2020we, wei2020implicit} suggest that searching for a flatter minima can improve the generalization ability of the model. To achieve this, we propose using Adversarial Weight Perturbation \cite{wu2020adversarial} to identify flat local minima of the empirical risk. The worst-case perturbations $\phi^{\ast}$  of the model weights which increases the loss dramatically is formulated as:
\begin{equation}
    \phi^{\ast} = \argmax_{\phi \in\mathcal{B}_{p}(\theta, \gamma)}\mathcal{L}(f_{\theta + \phi}(x), y),
\end{equation}
 where $\mathcal{B}_p(\theta, \gamma) = \{\upsilon \in \Theta : \parallel \theta - \upsilon \parallel _ {p} \leq \gamma\}$ is the feasible region of any perturbation $\phi$. The AWP adds the worst-case perturbation to the model weight, so that $\mathcal{L}(f_{\theta + \phi^{\ast}}(x), y)$ becomes the supremum value in $\mathcal{B}_p(\theta, \gamma)$.  Therefore, optimizing $\mathcal{L}(f_{\theta + \phi^{\ast}}(x), y)$ pushes $\theta$ adjust its values such that the loss landscape is more flatten with the same capability $\mathcal{B}_p(\theta, \gamma)$. As a result, $f$ become more stable under input image's changes.

Similar to adversarial example perturbation \cite{madry2017towards}, $\phi^{\ast}$ is generated by projected gradient method as follows:
\begin{equation}
    \phi^{\ast} \leftarrow \Pi_{\theta}^{\gamma} \left( \phi + \eta \frac{\nabla \mathcal{L}(f_{\theta + \phi}(x), y)}{ \parallel \nabla \mathcal{L}(f_{\theta + \phi}(x), y) \parallel} \parallel \theta\parallel\right ),
\label{eqn:phi_ast}
\end{equation}
 where $\Pi_{\theta}^{\gamma}$ is an operator that projects its input into the feasible region ${B}_{p}(\theta, \gamma)$, and $\eta \in \mathbb{R}$ is the step size. In fact, we empirically find that using a \textbf{one-step} projection for $\phi^{\ast}$ is sufficient for the model's robustness under the image corruptions formed by compression. By adding $\phi^{\ast}$ in Eq. \ref{eqn:phi_ast} to $\theta$, it is straightforward to bound $\mathcal{L}(f(x_c),y)$ in Eq. \ref{eqn:theorem1}.  

\subsubsection{Intra-model collaborative learning}
\textbf{Hilbert-Schmidt Independence Criterion (HSIC)}. Let $\mathcal{T}$ and $\mathcal{G}$ be two separable Reproducing Kernel Hilbert Spaces (RKHS) on metric spaces $\mathcal{U}$ and $\mathcal{V}$, respectively. HSIC measures the {dependency} between two random variables $U$ and $V$ {from} a {joint distribution} on $\mathcal{U}$ and $\mathcal{V}$, by evaluating the cross-covariance of the nonlinear transformations of the two random variables:
\begin{equation}
    HSIC(U, V) =  \parallel \mathbb{E}[\zeta(U)\psi(V)^T]  - \mathbb{E}\zeta(U)\mathbb{E}\psi(V)^T \parallel _{HS}^{2},
\label{eqn: hsic_org}
\end{equation}
where $\parallel\cdot\parallel_{HS}$ is the Hilbert-Schmidt norm, which becomes the Frobenius norm in finite dimensions. And, $\zeta:\mathcal{U \rightarrow \mathcal{T}}$ and $\psi: \mathcal{V} \rightarrow \mathcal{G}$ are nonlinear mapping functions. 
% that satisfy the \hl{reproducing property?} \simon{unclear} of $\mathcal{T}$ and $\mathcal{G}$, respectively. 
With appropriate transformation  $\zeta$ and $\psi$, HSIC is a dependence  test, which can identify the \textit{nonlinear dependencies} between $U$ and $V$ as follows: $HSIC(U, V) = 0 \Leftrightarrow U \perp V$. 

Also, inner products in  $\mathcal{T}$ and $\mathcal{G}$ are formed by positive definite kernel functions: $k(u, u^\prime) = \langle\zeta(u), \zeta(u^\prime)  \rangle _ {\mathcal{T}}$ and $l(v, v^\prime) = \langle\psi(v), \psi(v^\prime)  \rangle _ {\mathcal{G}}$. And, let $(U^\prime, V^\prime)$ and $(U^{\prime\prime}, V^{\prime\prime})$ be independent copies of $(U, V)$, then Eq. \ref{eqn: hsic_org} can be expressed as follows:
\begin{equation}
\begin{split}
    &HSIC(U, V) =  \mathbb{E}[k(U, U^\prime)l(V, V^\prime)] \\&- 2\mathbb{E}[k(U, U^\prime)]\mathbb{E}[l(V, V^{\prime\prime})] +  \mathbb{E}[k(U, U^\prime)]\mathbb{E}[l(V, V^\prime)].
\end{split}
\end{equation}
\textbf{Estimation of HSIC. }The empirical estimation of HSIC with an bias of $\mathcal{O}(\frac{1}{n})$ using $n$ samples \textit{i.i.d} drawn $\{(u_i, v_i) \}_{i=1}^{n}$ from the joint distribution $(U, V)$ is provided as follows \cite{gretton2005measuring}:
\begin{equation}
    \widehat{HSIC}(U, V) = \frac{1}{(n-1)^2}\text{\textbf{tr}}(KHLH),
\end{equation}
where $K_{i,j}=k(u_i, u_j)$, and $L_{i,j} = l(v_i, v_j)$ are kernel matrices for the kernels $k$, and $l$, {respectively,} and $H_{i,j}=\delta_{i,j}-\frac{1}{n}$ is a centering matrix. Regarding the kernel functions, Theorem 4 in \cite{gretton2005measuring} suggested that an universal kernel, such as Laplace and Gaussian RBF kernel, can guarantee the HSIC to detect any dependency between $U$ and $V$.

\textbf{HSIC for maximizing the geometrical similarity}. Let $\tau$ and $\rho$ be two different qualities of deepfakes, (e.g., raw vs. compressed). Consider the $l$-th layer of the learning network $f$, we denote the learning features of a mini-batch of $B$ images from $\tau$ and $\rho$ are $Z_{l}^{\tau}=\{u_i\}_B$, {and} $Z_{l}^{\rho} =\{v_i\}_B$, respectively,  where $u_i, v_i \in \mathbb{R} ^{H\times W \times C}$ and $H, W$ and $C$ are the height, width, and channel number. Our regularization aims to maximize the dependency between $Z_{l}^{\tau}$ and $Z_{l}^{\rho}$ via a mini-batch of representations. In other words, we try to minimize the following loss:
\begin{equation}
    \mathcal{L}_{col}(\tau,\rho) = -\sum_{l \in L}\widehat{HSIC}(Z^{\tau}_{l},Z_{l}^{\rho}),
\label{eqn:hsic_col}
\end{equation}
where $L$ is a predetermined collection of layers to apply the collaborative loss. And, the computational complexity for calculating Eq. \ref{eqn:hsic_col} is $\mathcal{O}(B^2L)$, which can be reduced to $\mathcal{O}(BL)$ when applying random Fourier features \cite{rahimi2007random}.

\begin{table*}[th!]
\centering
\resizebox{.80\textwidth}{!}{%
    \begin{tabular}{l!{\color{black}\VRule[1pt]}ccccccc!{\color{black}\VRule[1pt]}c}
    \hline
     \multirow{2}{*}{Model}  & \multicolumn{8}{c}{Test Set AUC (\%)} \\
     \cmidrule(lr){2-9}
      &  {NT} & {DF} & {F2F} & {FS} & {FSH}  & CDFv2 & FFIW10K & \textit{Avg}\\ 

    \hline \hline
 
    \multicolumn{9}{c}{\textit{\textbf{Video Compression (raw + c23 + c40 of test set)}}}\\
    \hline
    MesoNet~\cite{afchar2018mesonet}$^\lozenge$   & 70.24   &93.72   &94.15   &85.17   &96.00   &80.52   &94.56   &\textit{87.77}\Tstrut\\
    % Dogonadze \textit{et al.}~\cite{dogonadze2020deep} & 75.75 & 95.50 & 91.90 & 93.46 & 94.09 &89.37 & 92.73 & \textit{90.04}\\
      R\"ossler \emph{et al.}~\cite{rossler2019faceforensics++} $^\lozenge$   & 89.64   &99.05   &97.89   &98.83   &98.50   &97.49   &\textbf{99.17}   &\textit{97.22}\\
     $F^{3}\text{Net}$~\cite{qian2020thinking} $^\lozenge$ &86.79   &98.73   &96.32   &97.82   &97.45   &95.06   &97.94   &\textit{95.73}\\
     MAT~\cite{zhao2021multi}$^\lozenge$ &86.79   &98.73   &96.32   &97.82   &97.45   &95.06   &97.94   &\textit{95.73}\\
     Fang \& Lin~\cite{fang2021intra}   &89.30   &98.98   &97.33   &98.43   &98.66   &96.58   &98.94   &\textit{96.89}\\
     \hline
     SBIs \cite{shiohara2022detecting}$^\dagger$ & 78.33 & 95.19 & 79.74 & 80.37 & 80.48 & - & - &\textit{82.82} \\
     \hline
    BZNet \cite{lee2022bznet}$^\dagger$ & 80.12 & 98.81 & 94.10 & 97.71 & - &- & - & 91.01\\
     ADD~\cite{ble2022add}$^\dagger$   &86.26 &96.23 &90.62 &95.57 &95.94  &- &-   & \textit{92.92}  \\
    % \hline
    % \ResNet50 &224  &88.17   &99.07   &98.32   &98.44   &98.63   &96.49   &98.81   &\textit{96.85}\\
    \specialrule{1pt}{0pt}{0pt}
    \rowcolor{backcolour}\SystemName-R (\textit{\textbf{ours}})  &{91.25}   &\textbf{99.54}   &{98.34}   &{99.01}   &{99.12}   &{98.36}   &{99.10}   &{\textit{97.82}}\Tstrut\\
    % \cdashlinelr{1-10}
    \rowcolor{backcolour}\SystemName-E (\textit{\textbf{ours}})&\textbf{94.92}   &{99.53}   &\textbf{98.94}   &\textbf{99.27}   &\textbf{99.12}   &\textbf{98.38}   &{99.16}   &\textit{\textbf{98.47}}\Bstrut\\
    \hline
    \hline
    
    \end{tabular}%
}
\caption{\textbf{Classification performance in the quality-agnostic setting with video compression of test set}. The methods are trained using one of three approaches: simultaneously with three modalities (raw + c23 + c40),  individually  with each of the three modalities, or with a mid-level of compression (c23) to prevent performance degradation resulting from lossy compression. In the inference phase,\textbf{\textit{ video compression}} is  applied to the input. The best results are highlighted in \textbf{bold}. $\dagger$ and $\lozenge$ indicate results were obtained from methods' pre-trained weights and published code, respectively.} 
\label{tab:qual_agnos}
% \vspace{-8pt}
\end{table*}

\begin{table*}[t!]
\centering
\resizebox{.80\textwidth}{!}{%
    \begin{tabular}{l!{\color{black}\VRule[1pt]}ccccccc!{\color{black}\VRule[1pt]}c}
    \hline
     \multirow{2}{*}{Model}  & \multicolumn{8}{c}{Test Set AUC (\%)} \\
     \cmidrule(lr){2-9}
      &  {NT} & {DF} & {F2F} & {FS} & {FSH}  & CDFv2 & FFIW10K & \textit{Avg}\\ 

    \hline \hline
    \multicolumn{9}{c}{\textit{\textbf{Random Image Compression (JPEG on raw of test set)}}}\\
    \hline
    MesoNet~\cite{afchar2018mesonet}$^\lozenge$ &
70.23   &92.02   &88.32   &82.60   &91.84   &81.12   &91.87   &\textit{85.43}\Tstrut\\
    % Dogonadze \textit{et al.}~\cite{dogonadze2020deep} &66.09	&\textbf{94.99}	&\textbf{88.28}	&93.64	&91.38	&91.13	&92.57	&88.30 \\
     R\"ossler \emph{et al.}~\cite{rossler2019faceforensics++}$^\lozenge$  &69.89   &98.62   &{94.97}   &96.66   &96.76   &96.98   &98.81   &\textit{93.24}
    \\
    $F^{3}\text{Net}$~\cite{qian2020thinking}$^\lozenge$     & 70.95   &97.89   &92.83   &96.34   &94.72   &95.44   &97.19   &\textit{92.19}
  \\
     MAT~\cite{zhao2021multi} $^\lozenge$   & 69.53   &{98.96}   &\textbf{95.53}   &97.99   &96.97   &98.21   &{98.91}   &\textit{93.73}
 \\
     Fang \& Lin~\cite{fang2021intra}   & {75.49}   &98.32   &94.63   &97.64   &97.28   &96.67   &98.39   &\textit{94.06}
\\
\hline
SBIs \cite{shiohara2022detecting}$^\dagger$ &77.75 &97.83 & 82.05  & 86.10 &85.42 &- &- & \textit{85.83}\\
\hline
    BZNet \cite{lee2022bznet}$^\dagger$ &\textbf{ 79.00}   &98.77    & 95.23 &97.92  & - &- & - &92.73 \\
     ADD~\cite{ble2022add}$^\dagger$     & 75.84 & 96.83 & 92.23 &95.24 &96.00  & - & - &\textit{91.23}\\
    \specialrule{1pt}{0pt}{0pt}
     \rowcolor{backcolour}\SystemName-R  (\textit{\textbf{ours}})   &75.18   &98.86   &93.72   &{98.52}   &{98.18}   &{98.51}   &\textbf{98.96}   &{\textit{94.56}}\\
      \rowcolor{backcolour}\SystemName-E  (\textit{\textbf{ours}})   &{76.27}   &\textbf{99.20}   &{94.44}   &\textbf{98.69}   &\textbf{98.60}   &\textbf{98.52}   &{98.86}   &\textit{\textbf{94.94}}
\Bstrut\\
    \hline
    \hline
    
    \end{tabular}%
}

\caption{\textbf{Classification performance in the quality-agnostic setting with image compression of test set}. The training approach resembles that of Table \ref{tab:qual_agnos}'s setting, however, in the inference phase, \textbf{\textit{random image compression}} is  applied to the input. } 
\label{tab:qual_agnos_img}
\vspace{-8pt}
\end{table*}

\begin{algorithm}[t!]
\caption{\SystemName: Quality-Agnostic Deepfake detection.}
\begin{algorithmic}[1]
\Require   DNN $f$
parameterized by $\theta$, training dataset $\mathcal{D}$ with  M quality modalities   $\Tau = \{r, c_1, ...c_{M-1} \}$. Learning rate $\alpha_l$ and mini-batch size of ${B}$. Model weight perturbation size $\gamma$, step size $\eta$, and the number of steps $K$. Layers of $f$ to apply $HSIC$ $L$.
    \While{not converged}
        \For{$ \text{mini-batch } (X = [X_{\tau}]_{\tau\in \Tau}, \textrm{ } Y)\in \mathcal{D}$}
            \State  \texttt{\# One-step AWP}
            \State $\mathcal{L}_1 = \mathcal{L}(f_{\theta+\phi}(X),Y)/MB$
            \State $ \phi\leftarrow \Pi_{\theta}^{\gamma} \left( \phi + \eta \frac{\nabla \mathcal{L}_1}{ \parallel \nabla \mathcal{L}_1 \parallel} \parallel \theta\parallel\right )$
           
            \State $\theta \leftarrow \theta + \phi$
            \State $\mathcal{L}_1  \leftarrow  \mathcal{L}(f_{\theta}(X),Y)/MB$
            \State  \texttt{\# intermediate reps.}
            \State $ [Z_{l}^{\tau}]_{l \in L}^{\tau\in \Tau}:= f_{\theta}(X)$
            \State  \texttt{\# HSIC:  dependence maximization}
            \State $ \mathcal{L}_2 \leftarrow \sum^{\tau \neq \rho}_{\tau, \rho \in \Tau}\mathcal{L}_{col}(\tau, \rho)$
            \State  \texttt{\# Overall loss Eq. (\textcolor{red}{10})}
            \State $\mathcal{L}_{\SystemName} \leftarrow \mathcal{L}_1 + \alpha \times \mathcal{L}_2$
            \State $\theta \gets \theta - \alpha_l \cdot \nabla _{\theta} \mathcal{L}_{\SystemName}$ 
            \State $\theta \gets \theta - \phi$
        \EndFor
    \EndWhile
\end{algorithmic}

\label{alg:qad}
\end{algorithm}
\subsection{End-to-end training loss}
Given a training mini-batch $B$ that include all $M$ quality modalities $\Tau = \{r, c_1, ...c_{M-1} \}$, the overall collaborative learning loss in our~\SystemName\ framework is formulated as:
\begin{equation}
\begin{split}
     \mathcal{L}_{\SystemName} = & \frac{1}{MB }\sum_{\tau\in \Tau, i\in B}  \mathcal{L_{\phi^{\ast}}}(x_{\tau,i}, y_i) + \alpha \sum^{\tau \neq \rho}_{\tau, \rho \in \Tau}\mathcal{L}_{col}(\tau, \rho),
\end{split}
\label{eqn:qad}
\end{equation}
where $\alpha$ is a hyper-parameter to balance contribution of each loss. It is worth noting that our~\SystemName\ training loss is parameter-free, and is not affected by the order of the modalities. Further, unlike other model-based collaborative learning \cite{song2018collaborative}, our~\SystemName\ does not derive any sub-models. In other words, it can be integrated with any backbone, \textit{i.e.}, \ResNet50, and introduces no extra computation at the inference time. Note that Theorem 1 still holds when replacing the classification loss $\mathcal{L}(f(x),y)$ with any cross-entropy based loss, since $\mathcal{L}(f(x),y)$ is bounded  by the cross-entropy loss. Finally, we present our end-to-end algorithm for optimizing Eq. \ref{eqn:qad} in Algorithm \ref{alg:qad}, and  its pictorial illustration in Fig. \ref{fig:main_arch}.

\section{Experimental Results}
\begin{table*}[!t]
\centering
\resizebox{0.92\textwidth}{!}{%
{\renewcommand{\arraystretch}{1.5}%
    \begin{tabular}{l!{\color{black}\VRule[1pt]}c!{\color{black}\VRule[1pt]} r!{\color{black}\VRule[1pt]}ccccccc!{\color{black}\VRule[1pt]}c}
    \hline
    \multirow{2}{*}{Method} & \multirow{2}{*}{{w/ prior infor.}} &\multirow{2}{*}{\texttt{\#params}} & \multicolumn{8}{c}{Test Set AUC (\%)} \\
     \cmidrule(lr){4-11}
    
    & & &  {NT} & {DF} & {F2F} & {FS} & {FSH}  & CDFv2 & FFIW10K & \textit{Avg}\\
    \hline \hline 
    
    BZNet \cite{lee2022bznet}$^\dagger$ [$\times 3$]& {\texttt{\textcolor{red}{\textbf{Y}}}}  &{22M $\times$ {3}} & 91.01 & 99.30 &96.90 &98.82 &- & - &- &96.51 \\
    ADD~\cite{ble2022add}$^\dagger$ [$\times 3$]& {\texttt{\textcolor{red}{\textbf{Y}}}}  &{23.5M $\times$ {3}} &89.08 & 99.25 & 96.53  &98.21 & 98.25 & - & - &96.26\\
    \hline
    \ResNet50 [$\times 3$]  & {\texttt{\textcolor{red}{\textbf{Y}}}} &{23.5M $\times$ {3}} &88.96   &99.26   &97.04   &98.63   &98.71   &97.09   &98.58   &\textit{96.90}\\
    \rowcolor{backcolour}\SystemName-R (\textit{\textbf{ours}}) & {\texttt{\textcolor{green}{\textbf{N}}}} &{23.5M $\times$ {1}} & {88.85}   &{99.42}   &{97.77}   &{98.83}   &{98.93}   &{97.56}   &{98.93}   &{\textit{97.18}}\\
     \hline 
    \EffNet-B1[$\times 3$]& {\texttt{\textcolor{red}{\textbf{Y}}}}  &{6.5M $\times$ 3} &87.63   &99.05   &96.72   &98.16   &97.95   &96.70   &98.54   &\textit{96.39}\\
    \rowcolor{backcolour}\SystemName-E (\textit{\textbf{ours}}) &{\texttt{\textcolor{green}{\textbf{N}}}} &\textbf{6.5M $\times$ {1}} &\textbf{92.25}   &\textbf{99.46}   &\textbf{98.30}   &\textbf{99.08}   &\textbf{98.90}   &\textbf{97.50}   &\textbf{99.01}   &\textit{\textbf{97.79}}\\
    \hline
\end{tabular}%
}}
\caption{\textbf{Classification performance in the quality-aware setting with video compression of test set}. Except for our model, each model is trained with three modalities: raw, c23, and c40, respectively (denoted $[\times3]$). In the inference phase, while our \SystemName\ uses \textit{\textbf{one single}} pre-trained model, other methods use their \textit{\textbf{corresponding}} pre-trained model (\textit{e.g.}, pre-trained \ResNet-50 on raw) to detect a given testing input (\textit{e.g.}, raw). Reported performances are averaged score of the three modalities.}
\label{tab:qual_aware}
% \vspace{-8pt}
\end{table*}
\subsection{Dataset and pre-processing}
For evaluating our proposed method, we experiment with \textit{seven} different popular deepfake benchmark datasets: NeuralTextures (NT)~\cite{thies2019deferred}, Deepfakes (DF)~\cite{deepfakes}, Face2Face (F2F)~\cite{thies2016face2face}, FaceSwap (FS)~\cite{faceswap},  FaceShifter (FSH)~\cite{li2019faceshifter}, CelebDFV2 (CDFv2)~\cite{Celeb_DF_cvpr20}, and Face Forensics in the Wild (FFIW10K)~\cite{Zhou_2021_CVPR}. Besides the raw version, these videos are also compressed into two types: medium (c23) and high (c40), utilizing the H.264 codec and constant rate quantization parameters of 23 and 40, respectively. These effectively result in different quality of deepfakes, and details of these datasets are provided in our Supp. Material.
% Ablation study of the input image size is provided in Sec. \ref{subsec:ablation}.  
% Following the settings by {SOTA methods} \cite{ble2022add}, each dataset is randomly partitioned into three sets: training, validation, and test, with 720, 140, and 140 videos, respectively. We choose 64 frames at random from each video to generate 92,160, 17,920, and 17,920 images for the training, validation, and test sets, respectively.
% The Dblib library \cite{king2009dlib} is utilized to determine the largest face in each frame and scale it to a $128 \times 128$ square picture. \binh{move to supplementary due to space limit}

\subsection{Experimental Settings}
The models are trained with the Adam optimizer \cite{kingma2014adam} with a learning rate of 2e-3, scheduled by one-cycle strategy \cite{smith2019super} in 32 epochs. We use a mini-batch size of 64. In every training epoch, the model is evaluated \textit{ten} times, and we save the best one based on the validation accuracy. 
% We also apply early stopping \cite{prechelt1998early}, when the training model does not improve after \textit{twenty} consecutive validation steps. 
Regarding the backbone network, we use the \ResNet-50 \cite{he2016deep} (\SystemName-R) and \EffNet-B1 \cite{tan2019efficientnet}  (\SystemName-E) with their default input size of $224\times 224$ and $240 \times 240$, respectively.  The backbone models utilize pre-trained weights from~\ImageNet~dataset \cite{5206848}. Our hyper-parameters settings
$\{\sigma=6, \alpha=0.004, \gamma=0.002 \}$ are obtained by fine-tuning on \ResNet50  with NeuralTextures dataset and are kept the same throughout all datasets, whereas that of \EffNet-B1 are $\{\sigma=6, \alpha=0.002, \gamma=0.006 \}$.
% The performance of all models are shown by  AUC metric and \textit{Accuracy} (ACC) score. 
% The experiments are carried out on a single GeForce RTX 3090 24GB GPU with Intel Xeon Gold 6230R CPU @ 2.10GHz.
\subsection{Results}
\label{sub:results}
This section reports the results of our~\SystemName~and other baselines under two scenarios: 1) \textit{quality-agnostic} setting, which represents no model has prior knowledge of the input images' quality, and 2) \textit{quality-aware} setting, which baselines are required to know inputs' quality information.

\subsubsection{Quality-agnostic models} We use the popular deepfake detection benchmark methods on our datasets: 1) \textbf{MesoNet} with Inception layer by \cite{afchar2018mesonet}, 
% 2) the approach by \cite{dogonadze2020deep}, which utilized Inception~\ResNet V1 pre-trained on the VGGFace2 dataset, 
2) \textbf{Xception} model proposed by ~\cite{rossler2019faceforensics++},
3) \textbf{$F^{3}\text{Net}$ }\cite{qian2020thinking}, 4) \textbf{MAT} \cite{zhao2021multi} - a multi-attention deepfake detector, 5) a deviation of the method proposed by \cite{fang2021intra} to the \textbf{\textit{instance-based} collaborative learning}, SBIs \cite{shiohara2022detecting} - a self-blended method using real image only during training,  7) \textbf{ADD}~\cite{ble2022add} - a knowledge distillation-based approach for detecting low-quality deepfakes, and 8) \textbf{BZNet} \cite{lee2022bznet} - a super-resolution approach for improving detection of low-quality deepfakes. Each method has a different training approach to defend against performance degradation caused by lossy compression. The first five methods are trained with a mixture of the three data quality types (\textit{raw+c23+c40}). SBIs is trained with the mid-level of video compression (\textit{c23}), which is commonly adopted in many works. Meanwhile, ADD and BZNet models are trained on \textit{raw}, \textit{c23}, and \textit{c40}, respectively; however, in the inference phase, they are blindly tested over the entire dataset without the prior knowledge of quality types, and we report their average performance. In the test set, we include both video compression and random JPEG image compression~\cite{buslaev2020albumentations}.

The results for the video compression  are presented in Table \ref{tab:qual_agnos},  where our~\SystemName\ outperforms other SOTA baselines across multiple benchmark datasets. Notably, we achieve a significant improvement in AUC score of up to 5.28\% for heavily compressed datasets, such as NeuralTextures (89.64\% vs 94.92\%). We also surpassed previous works on various deepfake datasets by $0.44\%$ to $1.05\%$ points, with the exception of Deepfakes and FFIW10K datasets, which are easy to detect even when compressed. Compared to the collaborative learning baseline by \cite{fang2021intra}, which is a comparative benchmark, our~\SystemName\ still gains a decent improvement on average, up to $0.93\%$ and $1.54\%$ points with ~\SystemName-R and ~\SystemName-E, respectively. Finally, our \SystemName-E models achieved the highest score on average, reaching $98.47\%$. 

Regarding the random image compression experiment, the results are provided in Table \ref{tab:qual_agnos_img}. Although BZNet is marginally outperform our model on face-reenactment deepfakes (NT and F2F), our method still achieves the best performance with the highest scores on \textit{five} over \textit{seven} datasets. On average, our enhancements show decent improvements, with margins of 0.5\% and 0.88\% of \SystemName-R and \SystemName-E, respectively, compared to the second-best competitor (Fang \& Lin).

\begin{figure}[t]
\centering
\includegraphics[width=7.0cm]{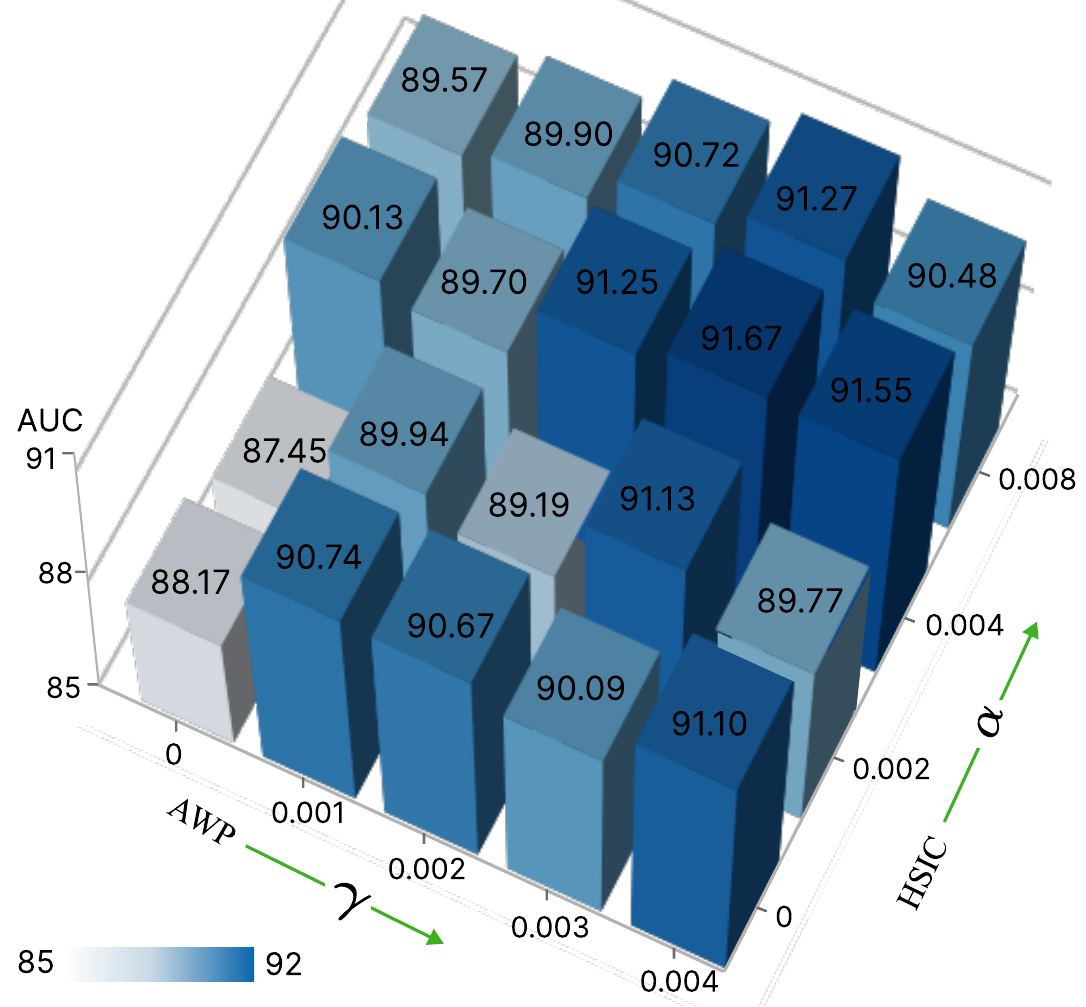}
\caption{Model's performance versus $\alpha$ and $\gamma$ on the NeuralTextures.}
\label{fig:sensitivity}
% \vspace{-8pt}
\end{figure}

\begin{table}[!t]
\centering
\resizebox{0.38\textwidth}{!}{%
\begin{tabular}{c!{\color{black}\VRule[1pt]} l !{\color{black}\VRule[1pt]} c   c }
        \hline
        \multicolumn{2}{c|}{\multirow{2}{*}{Model / loss}} & \multicolumn{2}{c}{\ResNet-50}\\
        \cmidrule(lr){3-4}
        \multicolumn{2}{c|}{} &  ACC (\%) & AUC (\%)  \tabularnewline
        \hline
        \hline
        \multicolumn{2}{c|}{Baseline}  & 78.8   & 88.2  \tabularnewline
        \hline
        \multirow{4}{*}{\textit{Coll. loss}} 
        & Soft-label & 77.0 & 84.0 \tabularnewline
        & Pairwise loss &79.7 &89.1 \tabularnewline
        & Center loss &79.8 &88.9 \tabularnewline
        &\cellcolor{backcolour} HSIC &\cellcolor{backcolour} \textbf{80.3} &\cellcolor{backcolour} \textbf{90.1}\tabularnewline 
        \hline
        \multirow{2}{*}{\textit{Adv. loss}}
        & AWP-KL & 80.9 & 89.4 \tabularnewline
        &\cellcolor{backcolour} AWP-XE &\cellcolor{backcolour} \textbf{81.7} &\cellcolor{backcolour} \textbf{90.7} \tabularnewline
        \specialrule{1pt}{0pt}{0pt}
        \rowcolor{backcolour}\multicolumn{2}{c|}{\SystemName~\textit{(ours)}}  & \textbf{82.2}   & \textbf{91.3} \tabularnewline
        \hline
        \end{tabular}%
}
        \caption{ Performance (ACC \& AUC) of \ResNet50 integrated  with different losses. } 
        \label{tab:ablation_loss}
\end{table}

\subsubsection{Quality-aware models} In this experiment, we compare our models with quality-aware benchmark baselines. In particular, beside ~\ResNet-50 and \EffNet-B1, for each of the \textit{raw}, \textit{c23}, and \textit{c40} datasets, we implement ADD  \cite{ble2022add} and BZNet \cite{lee2022bznet} models. Since ADD and BZNet are the best {performing methods}, in which they utilized knowledge distillation and super-resolution approaches, respectively, for detecting deepfakes in different qualities. Hence, we only include them in this experiment. In the inference phase, the performance of these models is validated with the prior knowledge of the input image's quality, \textit{i.e.}, \textit{c40} images are evaluated by the same quality \textit{c40} pre-trained models. Meanwhile, our universal~\SystemName\  is \textbf{blindly} evaluated without such prior knowledge. We integrate our~\SystemName\ on \ResNet50 and \EffNet-B1 and present their performance in Table \ref{tab:qual_aware}. As we can obbserve, our~\SystemName-E model performs slightly better or on par with \ResNet-50, BZNet, and ADD models, despite having only \textbf{one-third of the number of parameters} and \textbf{no prior knowledge of input image quality}. Moreover, when integrating with ~\EffNet-B1, \SystemName-E achieves a new SOTA performance with an improvement of up to $0.89\%$ points (97.79\% vs. 96.90\%), with a modest number of parameters (6.5M).

% \begin{figure}
% \centering
% \includegraphics[width=6.4cm]{Figures/nt_finetune_auc.pdf}
% \caption{Accuracy versus $\alpha$ and $\gamma$ on the NeuralTextures.}
% \label{fig:sensitivity}
% \vspace{-8pt}
% \end{figure}

\subsection{Ablation studies}
\label{subsec:ablation}
 \subsubsection{$\alpha$ and $\gamma$ of our loss} We investigate the sensitivities of our~\SystemName\ with respect to $\alpha$ and $\gamma$, and summarize the results of our analysis in Fig. \ref{fig:sensitivity}. In this study, we experiment with~\ResNet50 on the NeuralTextures dataset, which is {the hardest dataset to detect when compressed}. And, we vary the values of the hyperparameters $\alpha \in \{0.002, 0.004, 0.008 \}$ and $\gamma \in \{0.001, 0.002, 0.003, 0.004 \}$, {where} the value at $(\alpha, \gamma)=(0.0, 0.0)$ indicate the baseline. The results suggest that when $\alpha$ is greater than $0.002$, the performance of our model is high and stable, surpassing current SOTA with any $\alpha$ greater than $0.002$. Additionally, increasing $\gamma$ generally improves performance. Note that, as we did not tune the hyper-parameters to optimize the test accuracy, Section \ref{sub:results}'s hyper-parameters are not the best, despite outperforming all current methods on the datasets.

% \begin{table}[t!]
% \centering
% \resizebox{0.46\textwidth}{!}{%
% {\renewcommand{\arraystretch}{1.5}%
% \begin{tabular}{c|   l | c   c }
%     \hline
%     \multicolumn{2}{c|}{\multirow{2}{*}{Model / loss}} & \multicolumn{2}{c}{\ResNet-50}\\
%     \cmidrule(lr){3-4}
%      \multicolumn{2}{c|}{} &  ACC (\%) & AUC (\%)  \tabularnewline
%     \hline
%     \hline
%     \multicolumn{2}{c|}{Baseline}  & 78.8   & 88.2  \tabularnewline
%     \hline
%     \multirow{4}{*}{\textit{Coll. loss}} 
%     & Soft-label & 77.0 & 84.0 \tabularnewline
%     & Pairwise loss &79.7 &89.1 \tabularnewline
%     & Center loss &79.8 &88.9 \tabularnewline
%    &\cellcolor{backcolour} HSIC &\cellcolor{backcolour} \textbf{80.3} &\cellcolor{backcolour} \textbf{90.1}\tabularnewline 
%     \hline
%      \multirow{2}{*}{\textit{Adv. loss}}
%      & AWP-KL & 80.9 & 89.4 \tabularnewline
%      &\cellcolor{backcolour} AWP-XE &\cellcolor{backcolour} \textbf{81.7} &\cellcolor{backcolour} \textbf{90.7} \tabularnewline
%      \hline
%     \rowcolor{backcolour}\multicolumn{2}{c|}{\SystemName~\textit{(ours)}}  & \textbf{82.2}   & \textbf{91.3} \tabularnewline
%     \hline
% \end{tabular}%
% }}
% \caption{Performance (ACC \& AUC) of \ResNet50 integrated  with different loss functions. } 
% \label{tab:ablation_loss}
% \vspace{-8pt}
% \end{table}

\begin{table}[!t]
\centering
\resizebox{0.46\textwidth}{!}{%
{\renewcommand{\arraystretch}{1.6}%
    \begin{tabular}{@{\hskip2pt}c !{\color{black}\VRule[1pt]}c @{\hskip9pt}  c @{\hskip11pt} c@{\hskip11pt}  c @{\hskip11pt} c @{\hskip6pt} c@{\hskip4pt}  c @{\hskip4pt}!{\color{black}\VRule[1pt]}c @{\hskip2pt}}
    \hline
    {\SystemName}&  {NT} & {DF} & {F2F} & {FS} &{FSH}  &CDFv2 &FFIW10K & \textit{Avg}\tabularnewline
     \hline
      \hline
    \multicolumn{8}{c}{\textbf{\ResNet18}} \tabularnewline
    \hline
    \xmark &88.73   &98.93   &98.05   &98.06   &98.67   &97.09 &98.72 & \textit{96.89}

 \tabularnewline
    \rowcolor{backcolour}\checkmark &\textbf{91.38}   &\textbf{99.32}   &\textbf{98.32}   &\textbf{99.19}   &\textbf{98.94} & \textbf{97.97}		&\textbf{99.10} & \textbf{\textit{97.75}}
  
 \tabularnewline
    \hline
    \multicolumn{8}{c}{\textbf{\ResNet34}} \tabularnewline
    \hline
    \xmark  & 88.26   &99.01   &97.98   &98.67   &98.88  &96.61	&98.97 &\textit{ 96.91}

\tabularnewline
    \rowcolor{backcolour}\checkmark & \textbf{92.87}   &\textbf{99.30}   &\textbf{98.37}   &\textbf{99.10}   &\textbf{99.21} & \textbf{98.48}	&\textbf{99.14} & \textit{\textbf{98.07}}

 \tabularnewline
    \hline
    \multicolumn{8}{c}{\textbf{\EffNet-B0}} 
\tabularnewline
    \hline
    \xmark & 86.12   &99.29   &{97.93}   &98.21   &98.42  
  & {97.81}	&{98.80} & \textit{96.65}
\tabularnewline
   \rowcolor{backcolour}\checkmark & \textbf{91.99}   &\textbf{99.33}   &\textbf{98.66}   &\textbf{99.15}   &\textbf{99.00}  &\textbf{98.38}	&\textbf{99.12} & \textit{\textbf{97.95}}
  \tabularnewline
   \hline
\end{tabular}%
}}
\caption{Performance (AUC) of \ResNet18, \ResNet34, and  \EffNet-B0 baseline and their integration with our~\SystemName\ training framework.} 
\label{tab:ablation_backbone}
% \vspace{-12pt}
\end{table}

 \subsubsection{Selection of losses} We study different alternatives for the \textit{collaborative learning loss} and the \textit{adversarial weight perturbation} approach. In particular, for the collaborative learning loss, we apply the loss function that was introduced by~\cite{song2018collaborative}, in which they aggregate the logits of different views, combining them with their true labels to generate the soft labels. Besides, we replace our HSIC regularization with intermediate pairwise loss (Eq. \ref{eqn:col}) and center loss. Regarding the {adversarial weight perturbation}, we further apply the KL divergence between the representations of raw and compressed images to perturb the model's weights. 

We report the results in Table \ref{tab:ablation_loss}, where we observe that the soft label loss fails to improve the baselines due to a lack of low-level representation agreement between different image qualities. While both pairwise and center loss slightly improve the baselines and are unstable with different model architecture, our HSIC consistently achieves the best performance by relaxing the instance-base constraint. Meanwhile, we can observe the model's performance drops in terms of both ACC and AUC, when replacing our AWP with cross-entropy loss with KL divergence. Generally, this experiment shows that using pairwise differences of various quality image representations at the output, such as soft label, pairwise constraint, or AWP-KL, for optimizing the model can hinder its convergence to the optimal parameters. 

 \subsubsection{Experiment with different backbones} Table 
\ref{tab:ablation_backbone} shows the comparability of our~\SystemName\ with \textit{three} different backbone networks: \ResNet18, \ResNet34, and \EffNet-B0. The hyperparameter settings are kept the same for \ResNet50 and \EffNet-B1. As shown in Table \ref{tab:ablation_backbone},  our~\SystemName\ consistently improves the baselines, from 0.86\% to 1.3\% points on average of seven deepfake datasets.

\begin{figure}[t]
\centering
\includegraphics[width=8.0cm]{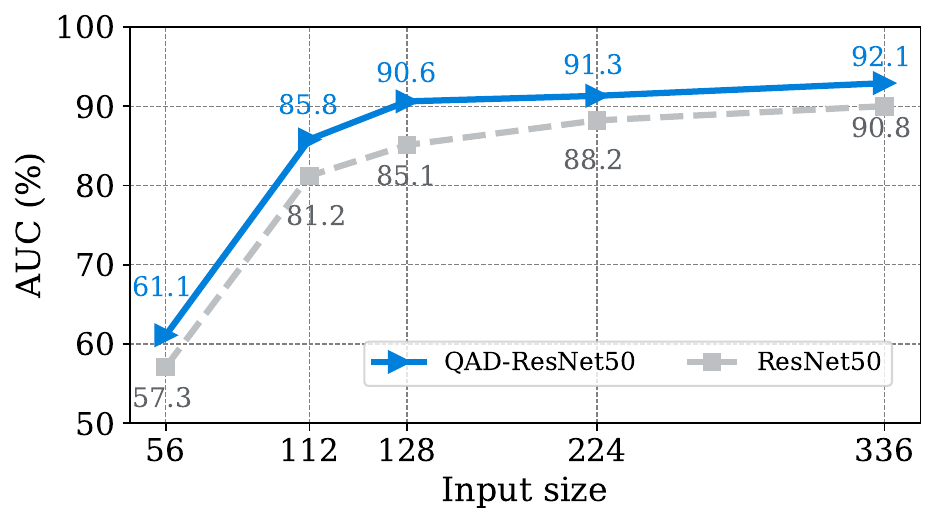}
\caption{Performance (AUC) of our proposed method at different input resolutions with NeuralTextures dataset.}
\label{fig:resolution}
% \vspace{pt}
\end{figure}

\begin{figure}[t]
\centering
\includegraphics[width=8.0cm]{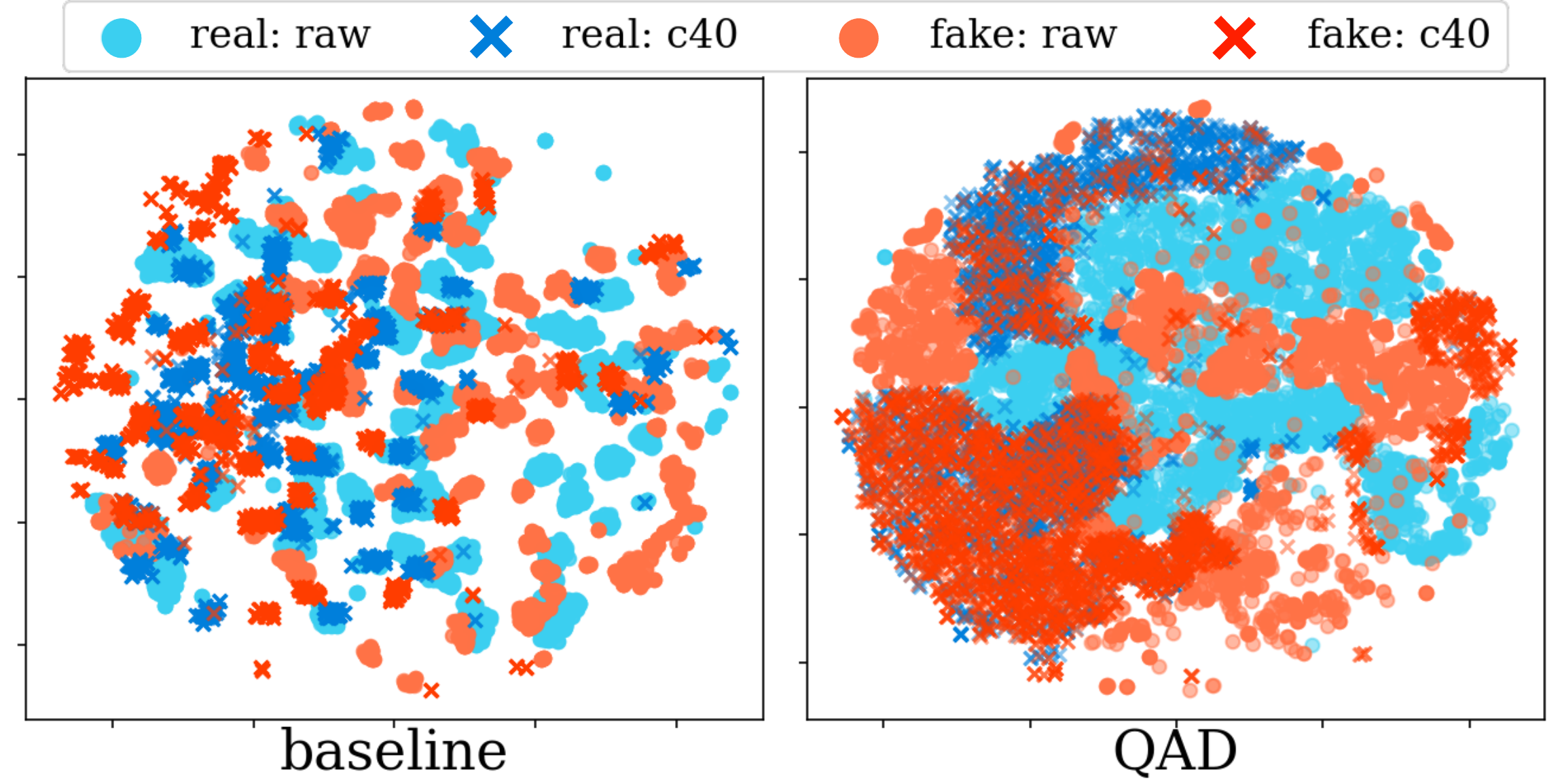}
\caption{ t-SNE visualisation of baseline  and our~\SystemName.}
\label{fig:tsne}
\vspace{-8pt}
\end{figure}
% \begin{figure}[t]
% \centering
% \includegraphics[width=7.5cm]{Figures/tSNE_NT.png}
% \caption{t-SNE visualisation of baseline (\textit{left}) and our~\SystemName\ (\textit{right}) with NeuralTextures dataset.}
% \label{fig:tsne}
% \vspace{-8pt}
% \end{figure}

\subsubsection{Performance at different input scales}  Unlike other classification tasks, a notable factor that substantially affects the deepfake detection performance, which is omitted by most previous works \cite{rossler2019faceforensics++, qian2020thinking, zhao2021multi}, is the input size of faces.  We resize the input images from $56$ to $336$ and demonstrate how it impacts our~\SystemName\ in comparison with \ResNet50 baseline. The experiment is performed with the NeuralTextures dataset, and its results are reported in Fig. \ref{fig:resolution}. We note that our proposed ~\SystemName\ and the baselines consistently improve their performance, when increasing the input size. Beside, our method also keep its staging improvement across the input resolutions compared to \ResNet50 baseline.

 \subsubsection{Feature distribution visualization} To verify the consistency of invariant representation upon the input quality, we draw the feature distribution of \EffNet-B1 and our~\SystemName-\EffNet-B1 pre-trained on NeuralTextures (with raw, c23, and c40 datasets) with {t-SNE} \cite{van2008visualizing}. The results are shown in Fig. \ref{fig:tsne}. As observed, our \SystemName\ model's representations are less dispersed both in terms of intra-class and inter-quality. This experiment demonstrates that traditional cross-entropy loss trained with multiple input quality  are confused due to the low-level constraints, while our~\SystemName\ enables the model to achieve more generalization regardless of input quality.

% \begin{table}[t!]
% \centering
% \resizebox{0.43\textwidth}{!}{%
% {\renewcommand{\arraystretch}{1.6}%
%     \begin{tabular}{c|  c | c  c | c c }
%     \Xhline{5\arrayrulewidth}
%     &\multicolumn{1}{c}{\multirow{2}{*}{\ResNet50}} & \multicolumn{2}{c}{\textit{collaborative loss}} & \multicolumn{2}{c}{\textit{adversarial loss}} \tabularnewline
%     \cmidrule(lr){3-4} \cmidrule(lr){5-6}
%     & &  HSIC & soft-label & AWP-XE & AWP-KL\tabularnewline
%     \hline
%     \hline
%     ACC & 76.3   &77.9   &75.5   &80.1   &76.4    \tabularnewline
%     AUC & 0.851   &0.872   &0.829   &0.890   &0.832   \tabularnewline
    
%     \Xhline{5\arrayrulewidth}
% \end{tabular}%
% }}
% \caption{Performance of \ResNet50 integrated  with different loss functions.} 
% \label{tab:ablation_loss}
% \end{table}

% \begin{figure}[t]
% \centering
% \includegraphics[width=6.5cm]{Figures/sensiti_NT.pdf}
% \caption{Ablation study on  $\alpha$ and $\gamma$. The blue and green lines indicate the results of our~\SystemName\ with different values of $\alpha$ and $\gamma$, respectively.}
% \label{fig:sensitivity}
% \end{figure}

% \begin{figure}[t]
% \centering
% \includegraphics[width=7.4cm]{Figures/tSNE_NT.pdf}
% \caption{t-SNE visualisation of baseline (\textit{left}) and ours~\SystemName\ (\textit{right}) with NeuralTexture dataset.}
% \label{fig:tsne}
% \end{figure}

\section{Conclusion}
Most deep learning-based deepfake detectors use a single model for each video quality, leaving an unsolved practical issue of their generalizability for detecting different quality {of} deepfakes. In this work, we propose a universal deepfake detection framework (\SystemName). Using intra-model collaborative learning, we minimize the geometrical differences of images in various qualities at different intermediate layers by the HSIC module. Moreover, our adversarial weight perturbation (AWP) module is directly applied to the model's parameters to provide its robustness against input image compression. Extensive experiments show that our~\SystemName\ achieves competitive detection accuracy and marks the new SOTA results on various deepfake datasets without prior knowledge of input image quality.

%\section*{Acknowledgments}
%This work was partly supported by Institute for Information \& communication Technology Planning \& evaluation (IITP) grants funded by the Korean government MSIT: (No. 2022-0-01199, Graduate School of Convergence Security at Sungkyunkwan University), (No. 2022-0-01045, Self-directed Multi-Modal Intelligence for solving unknown, open domain problems), (No. 2022-0-00688, AI Platform to Fully Adapt and Reflect Privacy-Policy Changes), (No. 2021-0-02068, Artificial Intelligence Innovation Hub), (No. 2019-0-00421, AI Graduate School Support Program at Sungkyunkwan University), and (No. RS-2023-00230337, Advanced and Proactive AI Platform Research and Development Against Malicious deepfakes).

{\small
\textbf{Acknowledgements.} This work was partly supported by Institute for Information \& communication Technology Planning \& evaluation (IITP) grants funded by the Korean government MSIT: (No. 2022-0-01199, Graduate School of Convergence Security at Sungkyunkwan University), (No. 2022-0-01045, Self-directed Multi-Modal Intelligence for solving unknown, open domain problems), (No. 2022-0-00688, AI Platform to Fully Adapt and Reflect Privacy-Policy Changes), (No. 2021-0-02068, Artificial Intelligence Innovation Hub), (No. 2019-0-00421, AI Graduate School Support Program at Sungkyunkwan University), and (No. RS-2023-00230337, Advanced and Proactive AI Platform Research and Development Against Malicious deepfakes).

\bibliographystyle{ieee_fullname}
\bibliography{egbib}
}

\begin{appendices}
\clearpage
% \onecolumn
\nocitesec{*}
% \begin{center}
% \vspace{30pt}\Large{\textbf{Supplementary Material}}\vspace{16pt} 
% \end{center}

In this supplementary material, we first provide a brief description of the datasets used in our experiment Section (Section \ref{sec:supp_data}). Next, the proof of Theorem 1 is provided in Section \ref{sec:supp_proof}. In Section \ref{sec:supp_expe},  we conduct ablation studies  about the detectors' robustness towards unseen corruptions.  Besides, we discuss the limitations of our proposed method in Section \ref{sec:supp_limit}. Finally, in Section \ref{sec:supp_vis}, we graphically illustrate the benefits of applying our \SystemName\ for deepfake detection problems.

\section{Datasets}
\label{sec:supp_data}
We describe here \textit{seven} popular benchmark deepfake datasets used to verify our proposed ~\SystemName:
\begin{itemize}
    \item \textbf{NeuralTextures}. Facial reenactment is a video re-rendering approach that uses the  \textit{Neural Textures}~\cite{thies2019deferred} technique. This method employs neural textures, which are learned feature maps placed on top of 3D model proxies, as well as a deferred neural renderer. The NeuralTextures dataset used in our study provide facial alterations to the mouth region, while the rest of the face remains unchanged. 
     \item \textbf{Deepfakes}. Each autoencoder in the DeepFakes dataset is trained on the source face and the target face separately before encoding the data. An artificial face is created by using the decoder trained on the target face to decode the embedding representation of the source face. Note that though DeepFakes originally referred to a particular face swapping technique, the term has now come to apply to AI-generated facial modification approaches in general.
    \item \textbf{Face2Face}. Face2Face~\cite{thies2016face2face} is a method of real-time facial recreation in which the identity of the target individual is maintained while their expression mimics that of the source. More specifically, a series of manually chosen key-frames is used in conjunction with a flexible model-based bundling strategy to recover the identification associated with the target face. Target backdrop and illumination are preserved while expression coefficients from the source face are transmitted.
    \item \textbf{FaceSwap}. FaceSwap~\cite{faceswap} is an easy-to-use program based on the visual architectures of the faces being swapped. To create a 3D representation of a person's face, 68 individual landmarks on the face are taken into account. Finally, it does color correction after projecting the facial areas back to the target face by reducing the pair-wise landmark errors.
    \item \textbf{FaceShifter}. FaceShifter~\cite{li2019faceshifter} is a two-step face-swapping system. The first step employs a generator with Adaptive Attentional Denormalization layers and an encoder-based multi-level feature extractor for a target face. In particular, the  Adaptive Attentional Denormalization  creates a synthetic face by fusing a person's identity with their physical characteristics. To improve face occlusions, they then created a unique Heuristic Error Acknowledging Refinement Network in the second phase.
    \item \textbf{CelebDFv2}. CelebDFv2 \cite{Celeb_DF_cvpr20} is a large, difficult dataset for deepfake forensics. It contains 590 original YouTube videos with subjects of various ages, ethnic backgrounds, and genders, as well as 5,639 DeepFake videos. The synthesized videos are generated by the Deepfake synthesis algorithm, followed by several refining steps targeting specific visual artifacts such as low resolution, color mismatch, and temporal flickering. 
    
    \item \textbf{Face Forensics In the Wild (FFIW10K)}. FFIW10K consists of 10,000 forgeries videos of high quality, with an average of three human faces every frame. Each video is created using one of three face-swapping techniques: DeepFaceLab \cite{perov2020deepfacelab}, FS-GAN \cite{nirkin2019fsgan}, and FaceSwap \cite{faceswap}, in order to increase the variety of manipulated videos.
\end{itemize}
For FaceForensics++ datasets, we follow the same preprocessing step as in ADD \cite{ble2022add} for
each modality. And, with a 3-tuple quality modality (raw, c23, and c40) in our experiment, we have 276,480, 53,760, and 53,750 for training, validation, and testing, respectively.
Similarly, for both CelebDF-v2 and FFIW10K datasets, we used 360,000, 49,200, and 49,200 images for training, validating, and testing, respectively.  To ensure the fair comparison, faces cropped from videos are resized to $128\times 128$, then prepossessed to the desired input size of each benchmark detector, \textit{e.g.}, $299 \times 299$ for XceptionNet. 

\section{Proof of Theorem 1}
\label{sec:supp_proof}

Let the optimization function be $\mathcal{L}(f(x), y)=1-\sigma_T(f(x,y))$, where $\sigma_T$ is the softmax function with temperature $T > 0$: 
\begin{equation}
    \sigma_T({f(x), y}) = \frac{\text{exp}(f(x,y) / T)}{\sum_{k=1}^{2}\text{exp}(f(x,k)/T)}.
\end{equation}
  We introduce a function class $\Phi_{\mathcal{W}} \subseteq [0,1]^{\mathcal{X}\times \mathcal{Y}}$ from the distribution of raw images:  $\Phi_{\mathcal{W}}= \{ (x_r, y) \mapsto\mathcal{L}(f(x_r),y): f\ \in \mathcal{F}\}$.

Let $\nu \in \{1,2, ..., \log_2(n) \}$ and $\tau_\nu  = 2 ^ {2-\nu }$, and we define function classes as follows:
\begin{equation}
\begin{split}
    \Phi _\nu = \{(x_c, y) \mapsto & \mathcal{L}(f(x_c), y) : \\ & \mathbb{E}_{\mathcal{D}} \left [| \sigma_T(f(x_r)) - \sigma_T(f(x_c)) | \right]\leq \tau_\nu  \}.
\end{split}
\end{equation}

For any $\mathcal{L} \in \Phi_{\nu}$, and $\delta \in (0,1)$, with probability at least $1-\delta$, the classical generalisation bound with the Rademacher complexity \cite{bartlett2002rademacher, mohri2012foundations} is defined as follows:
\begin{equation}
\begin{split}
    \mathbb{E}[\mathcal{L}(f(x_c), y)] \leq \mathbb{E}_{\mathcal{D}}[\mathcal{L}(f(x_c), y)] & + 2\mathfrak{R}_{\mathcal{D}}(\Phi_{\nu}) \\&+ \mathcal{O}\left(\sqrt{\frac{\log(2/\delta)}{2n}}\right),
\end{split}
\label{eqn:gen_bound}
\end{equation}
where
\begin{equation}
    \mathfrak{R}_{\mathcal{D}}(\Phi_{\nu}) = \frac{1}{n}\mathbb{E}_{\pi}\left [ \sup_{\mathcal{L} \in \Phi _\nu}\sum _{i=1}^{n}\pi_{i}\mathcal{L}(f(x_c), y)\right ],
\end{equation}
and $\pi_1, ..., \pi_n $ are i.i.d. Rademacher random variables with $P(\pi_i=1) = P(\pi_i=-1) = \frac{1}{2}$.
The Rademacher complexity $\mathfrak{R}_{\mathcal{D}}$ measures the rate that the empirical risk converges to the population risk.

Moreover, we also have:
\begin{equation}
\begin{split}
    \mathfrak{R}_{\mathcal{D}}(\Phi_{\nu}) &=  \frac{1}{n}\mathbb{E}_{\pi}\left [ \sup_{\mathcal{L} \in \Phi _\nu}\sum _{i=1}^{n}\pi_{i}\mathcal{L}(f(x_c), y)\right ] \\
    & = \frac{1}{n}\mathbb{E}_{\pi}\biggl [ \sup_{\mathcal{L} \in \Phi _\nu}\sum _{i=1}^{n}\pi_{i} ( \mathcal{L}(f(x_c), y) - \mathcal{L}(f(x_r), y) \\& \;\;\;\;\;\;\;\;\;\;\;\;\;\;\;\;\;\;\;\;\;\;\;\;\;\;\;\;\;\;\;\;\;\;\;\;\;\;\;\;\;\;\;\;\;\;\;\; +  \mathcal{L}(f(x_r), y))\biggr ] \\
    & \leq \frac{1}{n}\mathbb{E}_{\pi}\left [  \sup_{\mathcal{L} \in \Phi _\nu}\sum _{i=1}^{n}|\pi_i| |\mathcal{L}(f(x_c), y) - \mathcal{L}(f(x_r), y)  | \right ] \\
    & + \frac{1}{n}\mathbb{E}_{\pi}\left [  \sup_{\mathcal{L} \in \Phi _\mathcal{W}}\sum _{i=1}^{n}\pi_i \mathcal{L}(f(x_r), y)   \right ] \\
    & \leq \tau_\nu + \mathfrak{R}_{\mathcal{D}}(\Phi_{\mathcal{W}}) 
\end{split}
\label{eqn:bound_rade}
\end{equation}

{Replacing} Eq. \ref{eqn:bound_rade} into Eq. \ref{eqn:gen_bound}, we have:
\begin{equation}
\begin{split}
    \mathbb{E}[\mathcal{L}(f(x_c), y)] \leq & \mathbb{E}_{\mathcal{D}}[\mathcal{L}(f(x_c), y)] + 2 \tau_\nu \\
    & + 2 \mathfrak{R}_{\mathcal{D}}(\Phi_{\mathcal{W}}) +  \mathcal{O}\left(\sqrt{\frac{\log(2/\delta)}{2n}}\right).
\end{split}
\label{eqn:bound_1}
\end{equation}

In addition, for every $\mathcal{L}(f(x_c), y))$, there always exists $\tau_\nu$, such that:
\begin{equation}
    \tau_\nu \geq \mathbb{E}_\mathcal{D}\left[\parallel \sigma_T(f(x_r)) - \sigma_T(f(x_c)) \parallel \right] \geq \frac{1}{2}\tau_\nu -2^{1-\log_2 (n)}.
\end{equation}
Then
\begin{equation}
\begin{split}
    \tau_\nu & \leq \frac{4}{n} + 2\mathbb{E}_\mathcal{D}\left[\parallel \sigma_T(f(x_r)) - \sigma_T(f(x_c)) \parallel \right] \\
\end{split}
\end{equation}
Now, Eq. \ref{eqn:bound_1} can be rewritten as:
\begin{equation}
\begin{split}
    \mathbb{E}[\mathcal{L}(f(x_c), y)] \leq & \mathbb{E}_{\mathcal{D}}[\mathcal{L}(f(x_c), y)] + \frac{8}{n} \\ & 4\mathbb{E}_\mathcal{D}\left[| \sigma_T(f(x_r)) - \sigma_T(f(x_c)) | \right] \\
    & + 2 \mathfrak{R}_{\mathcal{D}}(\Phi_{\mathcal{W}}) +  \mathcal{O}\left(\sqrt{\frac{\log(2/\delta)}{2n}}\right).
\end{split}
\label{eqn:bound_2}
\end{equation}

Next, we rewrite the loss function for a compressed image, $x_c$, as follows:
\begin{align}
        & \mathcal{L}(f(x_c), y)  = \frac{\sum_{i\neq y}\exp(f(x_c, i) / T)}{\sum_{i=1}^{2}\exp(f(x_c, i) / T)} \nonumber \\
        & = \frac{1}{1+\frac{\exp(f(x_c, y) / T)}{\sum_{i\neq y}\exp(f(x_c, i) / T)}} \nonumber\\
        &  = \frac{1}{1+\exp(f(x_c, y) / T - \ln\bigl(\sum_{i\neq y}\exp(f(x_c, i)/T)))} \nonumber\\
        & = s \Bigl( -f(x_c,y)/T + \ln \bigl( \sum_{i\neq y} \exp \left ( f(x_c, i)/T \right ) \bigr)\Bigr),
\label{eqn:loss_derive}
\end{align}
where $s(\cdot)$ is the sigmoid function. However, note that the second term in the sigmoid function of Eq. \ref{eqn:loss_derive} belongs to the family of Log-Sum-Exp (LSE) function, and $s(\cdot)$  is a monotonically increasing function. Then, we have:
\begin{equation}
        \mathcal{L}(f(x_c), y) \geq s \Bigl( -f(x_c,y)/T +  f(x_c, \tilde{y})/T   \Bigr),
\end{equation}
where $\tilde{y}=  \argmax_{i\neq y}f(x_c, i)$.
Since we always have $s(t) \geq \frac{1}{2}\mathbb{I}(t \geq 0) , \forall t \in \mathbb{R}$, then
\begin{equation}
\begin{split}
    s \Bigl( -f(x_c,y)/T + & f(x_c, \tilde{y})/T   \Bigr) \geq \\ & \frac{1}{2} \mathbb{I} \Bigl( -f(x_c,y)/T +  f(x_c, \tilde{y})/T \geq 0  \Bigr),
\end{split}
\end{equation}
or
\begin{equation}
\begin{split}
    2\mathcal{L}(f(x_c), y) & \geq \mathbb{I} \Bigl( f(x_c,y)/T  \leq  f(x_c, \tilde{y})/T  \Bigr)\\
     & =  \mathbb{I} \Bigl( \hat{y}(x_c) \neq y \Bigr).
\end{split}
\label{eqn:bound_id}
\end{equation}
Combining Eq. \ref{eqn:bound_2} and Eq. \ref{eqn:bound_id},
    \begin{align}
        &\mathbb{E}\left[ \mathbb{I} \{\hat{y}(x_c)\neq y\ \}\right] \leq 2\mathbb{E}[\mathcal{L}(f(x_c), y)]  \nonumber\\ & \leq 2\mathbb{E}_{\mathcal{D}}[\mathcal{L}(f(x_c), y)] + \frac{16}{n} \nonumber\\ & 8\mathbb{E}_\mathcal{D}\left[\parallel \sigma_T(f(x_r)) - \sigma_T(f(x_c)) \parallel \right] \nonumber\\
    & + 4 \mathfrak{R}_{\mathcal{D}}(\Phi_{\mathcal{W}}) +  \mathcal{O}\left(\sqrt{\frac{\log(2/\delta)}{2n}}\right). 
    \end{align}

As softmax is a $L-Lipschitz$ function \cite{gao2017properties} with $L=1/T$, we obtain:

\begin{align}
    \mathbb{E}\left[ \mathbb{I} \{\hat{y}(x_c)\neq y\ \}\right] & \leq 2\mathbb{E}_{\mathcal{D}}[\mathcal{L}(f(x_c), y)] + \frac{16}{n} \nonumber \\  & \frac{8}{T}\mathbb{E}_\mathcal{D}\left[\parallel f(x_r) - f(x_c) \parallel \right] \nonumber\\ 
& + 4 \mathfrak{R}_{\mathcal{D}}(\Phi_{\mathcal{W}}) +  \mathcal{O}\left(\sqrt{\frac{\log(2/\delta)}{2n}}\right). 
\end{align}

\section{Additional Experiments}
\label{sec:supp_expe}

% \textbf{Details of the quality-aware deepfake detection experiment.} Table \ref{tab:qual_aware_details} provides the performance details of each method of the quality-aware deepfake detection experiments in our paper.

\begin{figure*}[t!]
\centering
\includegraphics[width=15.0cm,height=10.4cm]{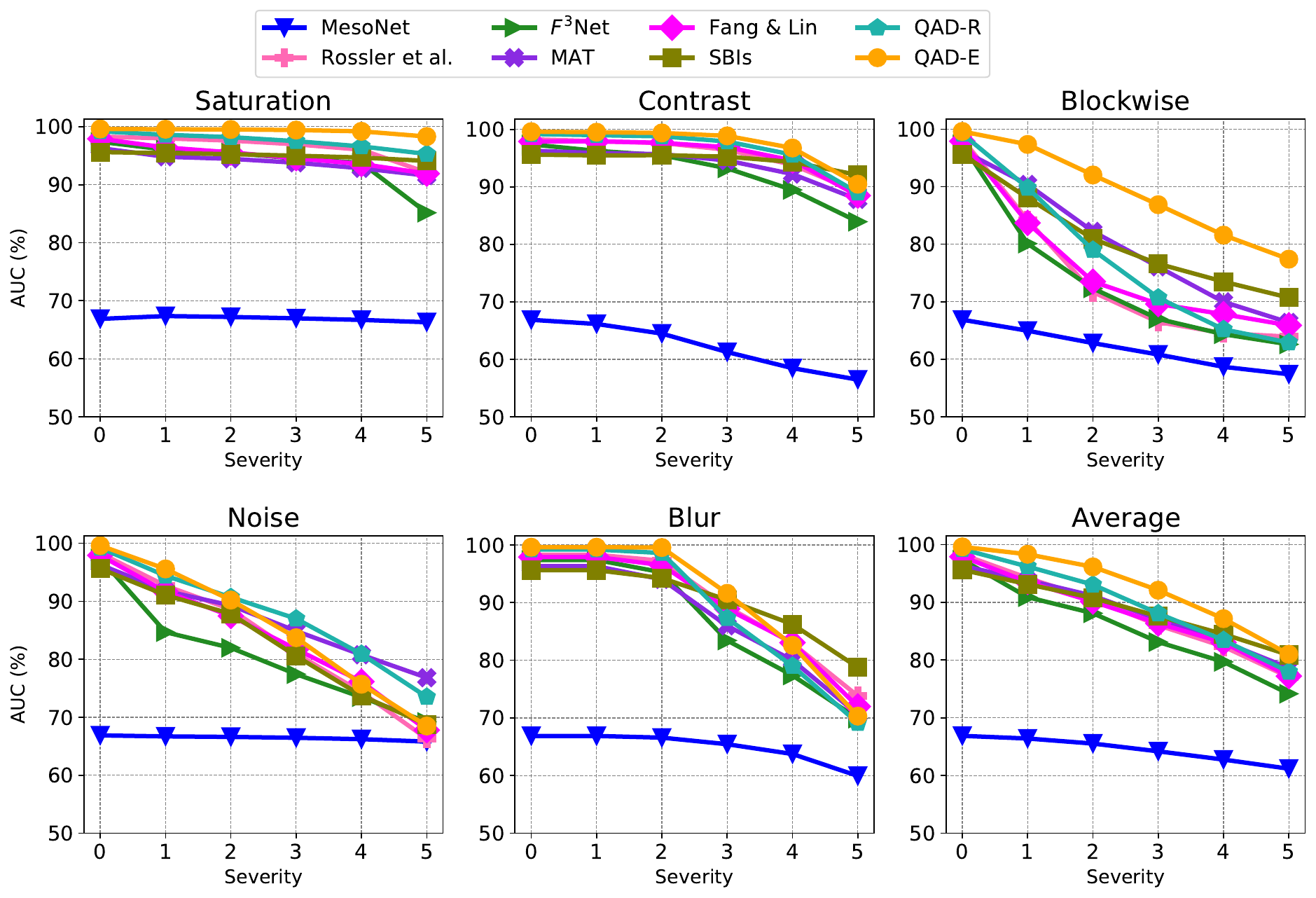}
\caption{Classification performance (AUC) of deepfake detectors under various corruptions with different severity levels.}
\label{fig:distortions}
\vspace{-8pt}
\end{figure*}

\textbf{Robustness against unseen corruptions.} Although our ~\SystemName\  does not intend to defend against all image corruption types, we investigate the robustness of our model and other detectors under unseen perturbations. All defenders in our experiment are trained on FaceForensics++  \cite{rossler2019faceforensics++} including ~\textit{five} typical deepfakes: NeuralTextures, DeepFakes, Face2Face, FaceSwap, and FaceShifter, with their quality modalities: raw, c23, and c40. In the inference phase, we apply \textit{five} operations with \textit{five} severity levels, as given in \cite{jiang2020deeperforensics}: saturation, contrast, block-wise distortion, white Gaussian noise, and blurring. The results are indicated in Fig. \ref{fig:distortions}. Although none of the perturbations are included in the training phase, generally, our proposed \SystemName-E achieves the best robustness compared to previous SoTA approaches. One may notice that MAT \cite{zhao2021multi} is a competitive defender; it obtains more robustness in the worst case by using a large input size of an image, \textit{i.e.}, $380 \times 380$, which can alleviate the perturbations' effects. Finally, we believe that our method can be generalized to different corruptions when including them in the training phase, making detectors more robust. However, this is out of our study's scope, which mainly targets deepfake compression.

\section{Limitations}
\label{sec:supp_limit}

We can point out two limitations of \SystemName. First, our proposed method relies on the existence of a $M$-tuple of quality modalities in the training dataset. While this requirement in the research environment is usually satisfied, a few deepfake datasets contain only videos in different qualities and conditions, such as DFDC \cite{dolhansky2020deepfake}. Therefore, possible future research could focus on utilizing unpaired images of various qualities to robust the deepfake detector.

Secondly, as we target detecting deepfake in multi-quality and we did not mining fine-grained deepfake artifacts as in \cite{zhao2021multi}, our ~\SystemName\ can be less generalized when validating across datasets. As shown in Table ~\ref{tab:cross_valid}, we trained all detectors on ~\textit{five} FaceForensics++ \cite{rossler2019faceforensics++} datasets with their \textit{three} versions: raw, c23, and c40. In the inference phase, we test the pre-trained models on DFDC \cite{dolhansky2020deepfake}  and WildDeepfake \cite{zi2020wilddeepfake} datasets. While our ~\SystemName, especially when integrating with \EffNet-B1, outperforms previous works on the WildDeepfake, it still needs to improve on the DFDC dataset. Our future work will focus on designing a good metric learning framework that can be generalized towards both input quality and cross-domain deepfakes.

\begin{table}[t!]
\centering
\resizebox{0.490\textwidth}{!}{%
{\renewcommand{\arraystretch}{1.6}%
    \begin{tabular}{l| c c | c c | c c}
    \hline
    \multirow{3}{*}{Method} & \multicolumn{2}{c|}{Training set} & \multicolumn{4}{c}{Test set}\\
     \cmidrule(lr){2-3} \cmidrule(lr){4-7} 
    &\multicolumn{2}{c|}{FF++} & \multicolumn{2}{c|}{DFDC} & \multicolumn{2}{c}{WildDeepfake} \tabularnewline
    \cmidrule(lr){2-3} \cmidrule(lr){4-5} \cmidrule(lr){6-7} 
    & ACC & {AUC} &   ACC & AUC &   ACC & AUC \tabularnewline
     \hline
    \hline
    MesoNet \cite{afchar2018mesonet} & 61.6   &65.6   &60.3   &71.4   &54.4   &55.8 
 \tabularnewline
    R\"ossler \emph{et al.} \cite{rossler2019faceforensics++}  &79.4   &86.4   &57.6   &66.0   &61.1   &66.4 \\
    $F^{3}\text{Net}$ \cite{qian2020thinking}    &75.4   &84.2   &54.1   &66.4   &58.9   &64.5  \\
     MAT  \cite{zhao2021multi}  &77.3   &86.8   &\textbf{65.1}   &\textbf{71.9}   &{63.6}   &{71.0} \\
     Fang \& Lin \cite{fang2021intra}   &80.7   &89.0   &63.2   &70.2   &63.1   &67.7\\
     % ADD [$\times 3$]     &75.9	&0.838	&56.5	&0.596	&58.8	&0.637
     SBIs \cite{shiohara2022detecting} &68.9 &86.0 &60.5 &70.9 &59.2 &65.5

\\
\hline
    % \cdashlinelr{1-8}
     % \ResNet50 \cite{he2016deep} & 78.8   &0.877   &60.9   &0.656   &62.0   &0.668 
  
    \SystemName-R  &{85.3}   &{93.4}   &61.4   &67.0   &62.5   &68.9\\ 
    \SystemName-E & \textbf{87.8}	&\textbf{95.6}	&56.5	&65.3	&\textbf{65.5}	&\textbf{74.7}\\
    \hline
\end{tabular}%
}}
\caption{Cross-validation performance of models that trained on FF++ and validated on DFDC and WildDeepfake datasets.} 
\label{tab:cross_valid}
% \vspace{-8pt}
\end{table}

\section{Grad-CAM Results} 
\label{sec:supp_vis}
\begin{figure*}[t]
\centering
\includegraphics[width=10.5cm]{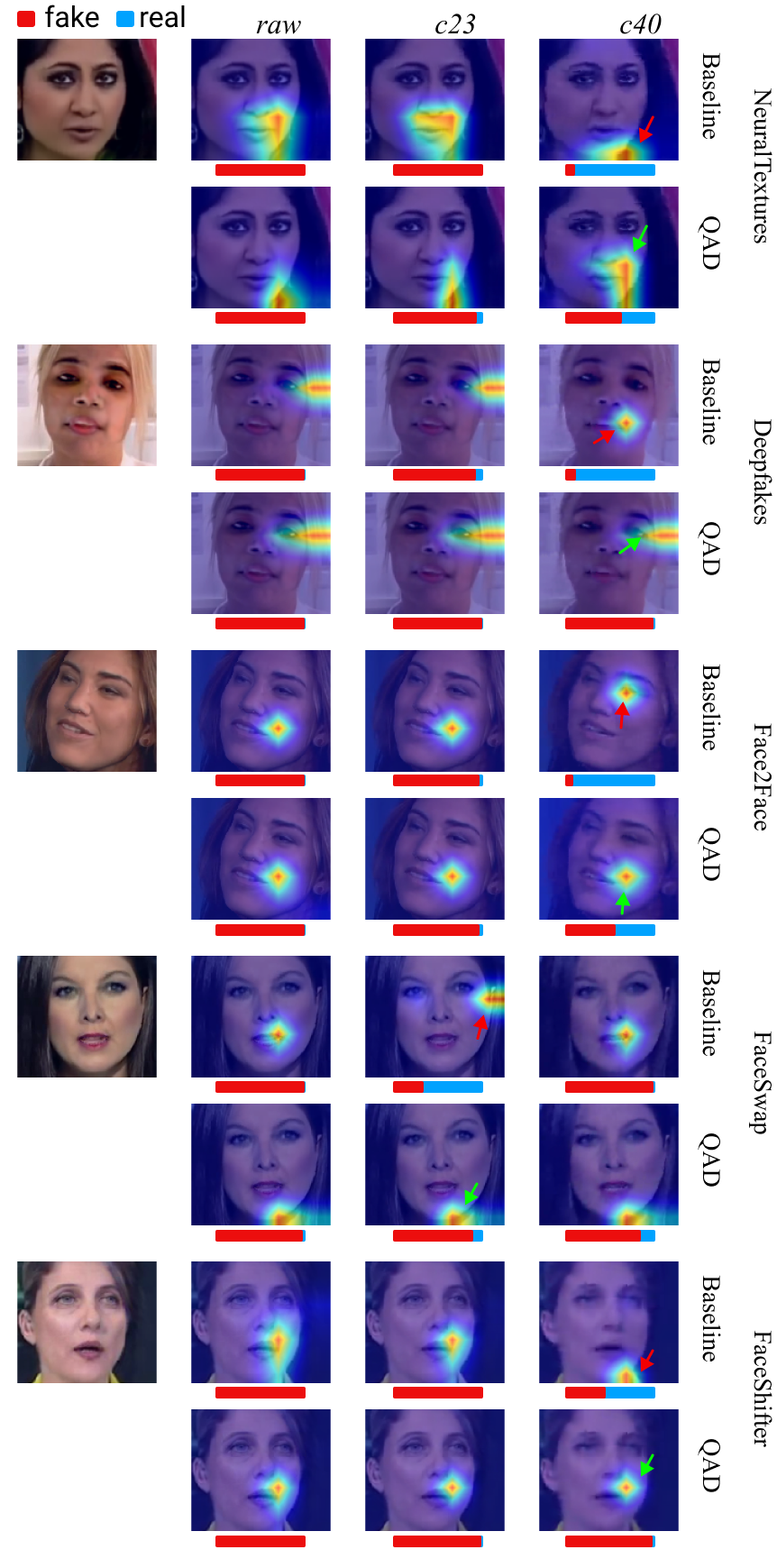}
\caption{Grad-CAM activation maps of \textbf{deepfake images} from NeuralTextures, DeepFakes, Face2Face, FaceSwap and FaceShifter dataset. The red arrows indicate the inconsistent activation regions created by the ~\ResNet50 baseline. The green arrows indicate the re-corrected activation regions created by our ~\SystemName\ framework. }
\label{fig:cam_confi}
\end{figure*}

\begin{figure*}[t]
\centering
\includegraphics[width=10.5cm]{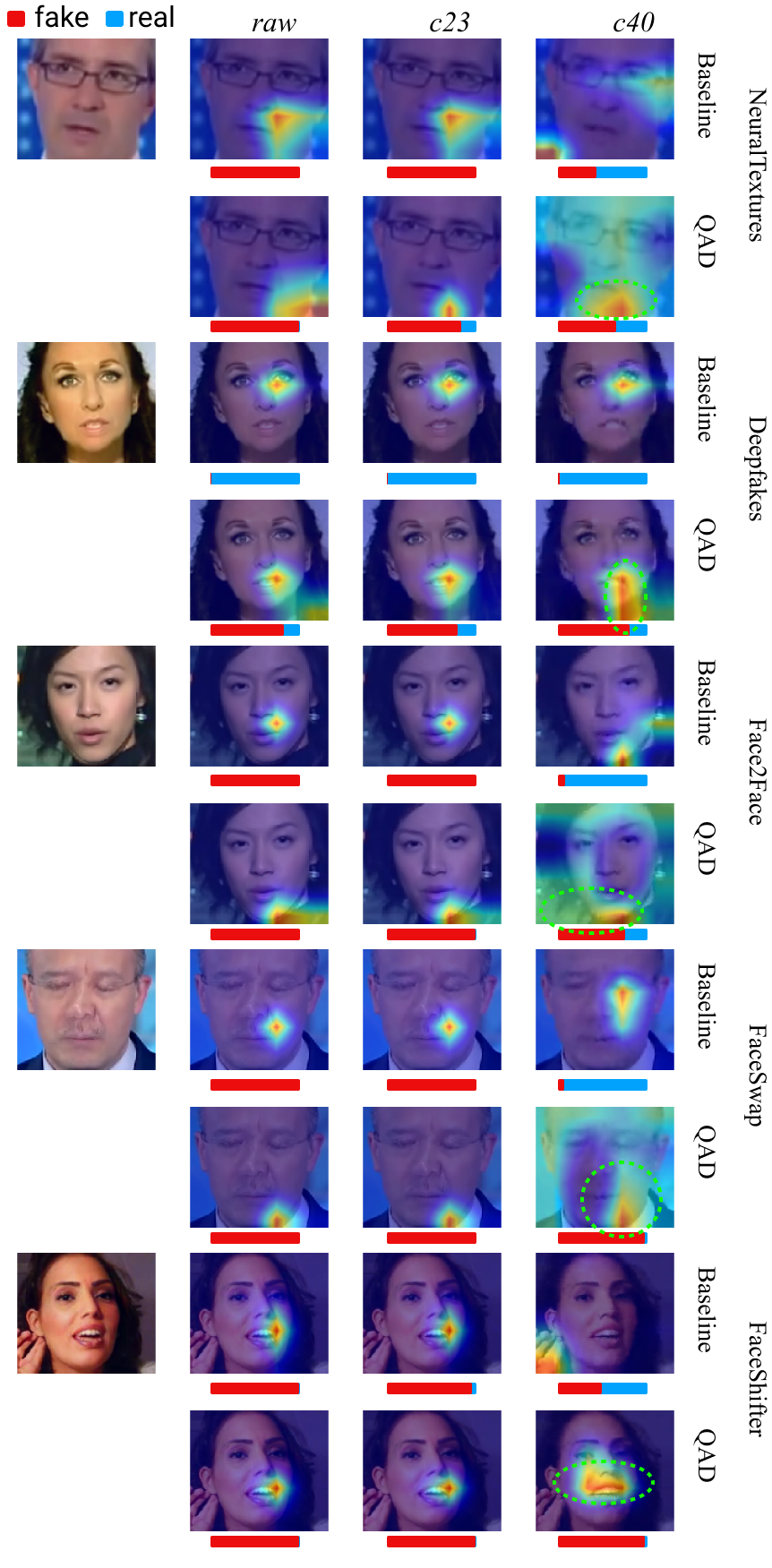}
\caption{Grad-CAM activation maps of \textbf{deepfake images} from NeuralTextures, DeepFakes, Face2Face, FaceSwap and FaceShifter dataset. In contrast with inconsistent or wrong activation regions from the baseline, our~\SystemName\ can enlarge the activation regions in the low-quality images and reconcile them with other quality modalities.}
\label{fig:cam_enl}
\end{figure*}

Gradient-weighted Class Activation Mapping (Grad-CAM) \cite{selvaraju2017grad} applies gradients from a high-probability prediction class to a lower level convolutions layer, resulting in a coarse localization map that highlights critical locations in the image. Positive layer outputs and higher gradient values generate more activation areas, which are illustrated in red in  Fig.~\ref{fig:cam_confi} and Fig. ~\ref{fig:cam_enl}. On the other hand, negative pixels or low gradients produce fewer activation regions, which are represented in blue. In this experiment, we visualize \ResNet50 baseline and our~\SystemName-R on the \textit{five} FaceForensics++ datasets: NeuralTextures, Deepfakes, Face2Face, FaceSwap, FaceShifter, and explain the benefits of our training frameworks as follows:
\begin{itemize}
    \item \textbf{Consistently activating important regions across quality modalities}. Since \ResNet-50  is trained without any regularization, it considers images of different quality as different images. Therefore, the activation regions can vary across quality modalities, which is indicated by the red arrows in Fig.~\ref{fig:cam_confi}, resulting in a large proportion of wrong predictions in low-quality images. In contrast, our ~\SystemName~ utilizes the HSIC to maximize the dependence between quality modalities, it regulates activation regions to be similar, ensuring its generalizability for detecting different quality deepfakes, as show in Fig.~\ref{fig:cam_confi}.
    \item \textbf{Expanding attention regions in low-quality images}. Under heavy compression, subtle differences and artifacts for distinguishing deepfakes can be {diminished}. As we can observe in Fig. \ref{fig:cam_enl},~\ResNet50 activates different regions, resulting in inconsistent prediction among quality modalities.  Meanwhile, our ~\SystemName\ by utilizing the AWP assists in enlarging activation regions on the low-quality images, which are indicated by the green circles in Fig. \ref{fig:cam_enl}. Therefore, it accumulates information from several regions inside the low-quality image to make an overall prediction, making the detector more accurate.
\end {itemize}

\end{appendices}

\end{document}